\documentclass{article}
\usepackage[final]{corl_2026}
\usepackage[utf8]{inputenc}
\usepackage[T1]{fontenc}
\usepackage{amssymb}
\usepackage{amsmath}
\usepackage{amsbsy}
\usepackage{amsfonts}
\usepackage[nice]{nicefrac}
\usepackage{breqn}
\usepackage{mathtools}
\usepackage[linesnumbered,ruled,vlined]{algorithm2e}
\usepackage{makecell}
\usepackage{graphicx}
\usepackage{subcaption}
\usepackage[font=small, labelfont=bf]{caption}
\usepackage{svg}
\usepackage{float}
\usepackage{dblfloatfix}
\usepackage{hyperref}
\usepackage{booktabs}
\usepackage{microtype}
\usepackage{multirow}
\usepackage{booktabs}
\usepackage{adjustbox}
\allowdisplaybreaks
\usepackage{needspace}
\usepackage{wrapfig}
\usepackage{algorithm2e}
\RestyleAlgo{ruled}
\usepackage{booktabs}
\usepackage{graphicx}
\usepackage{capt-of} 

\usepackage{xcolor}
\usepackage{enumitem}
\newcommand{\todo}[1]{\textcolor{red}{TODO: #1}}
\newcommand*{\algname}{\textsc{ASD}\xspace}
\newcommand{\algline}[1]{\ref{#1}}
\definecolor{skyblue}{RGB}{95,190,230}
\title{Training Generalist Navigation Policies via \\ Action Self-Distillation}

\usepackage{listings}
\usepackage{xcolor}
\author{
  Bowen Feng\textsuperscript{\dag} \quad Keqin Wang\textsuperscript{\dag} \quad Yulong Yang\textsuperscript{\dag} \quad Amogh Joshi\\
  \textbf{Christine Allen-Blanchette} \quad \textbf{Dhruv Shah}\textsuperscript{$\ddagger$} \quad \textbf{Felix Heide}\textsuperscript{$\ddagger$}\\
  Princeton University, USA\\
  \texttt{$\{$bf2192, kw4335, yy0555, aj0699$\}$@princeton.edu}\\
  \texttt{$\{$ca15, shahd, fheide$\}$@princeton.edu}\\[4pt]
  \href{https://asd-nav.github.io}{\textbf{\textcolor{skyblue}{asd-nav.github.io}}
  }
}

\begin{document}
\maketitle
{\let\thefootnote\relax%
 \footnotetext{\textsuperscript{\dag}Equal contribution. \quad \textsuperscript{$\ddagger$}Joint supervision.}}

\begin{abstract}
Language provides a natural interface for humans to specify and reason about goals, but training language-conditioned robotic policies and reasoning typically requires expensive manual annotations. We propose \textit{Action Self-Distillation} (\algname{}), a method for learning language-conditioned reasoning for robotic navigation without human-annotated language or reasoning labels. \algname{} uses an open-source VLM, which contains broad world knowledge and reasoning capabilities, as \emph{both} a teacher network and a student policy. During training, the self-teacher receives privileged trajectory feedback and infers the underlying navigation goal together with goal-oriented reasoning. The student is then trained to condition on the inferred navigation goal, predict future ego-frame trajectory, and provide a description of the scene semantics, plan, and reasoning, from onboard observations alone. Across 13 challenging indoor and outdoor real-world navigation tasks, our method improves success rate over the strongest baseline by $30\%$ in indoor settings, $30\%$ on multi-stage tasks, and achieves $95\%$ success rate on outdoor tasks, validating that privileged trajectory feedback can enable self-distillation for scalable generalist navigation policies without human annotations.
\end{abstract}



\section{Introduction}\label{sec:intro}
Vision-language-action (VLA) models have emerged as a successful approach to robot learning, enabling policies to condition on natural-language commands while grounding them in visual observations~\citep{brohan2023rt2, kim2024openvla}. For navigation, this paradigm is especially appealing because language provides a deployable interface for specifying goals in human-centered environments, without requiring privileged map coordinates or goal images~\citep{shah2023lm, hirose24lelan, zhou2024navgpt, zhou2024navgpt2}. However, training VLMs to end-to-end navigation typically requires datasets with goal descriptions, scene semantics or reasoning traces, followed by supervised fine-tuning~\citep{hirose2025omnivla, hirose24lelan}. These annotations are expensive to obtain at scale, and many widely used robot navigation datasets contain only sensor observations and trajectories, with no language supervision~\citep{shah2021rapid, frodobots2k, hirose24lelan, shah2023gnm, shah2023vint, sridhar2024nomad}.
\begin{figure}[t]
    \centering
    \includegraphics[width=0.7\linewidth]{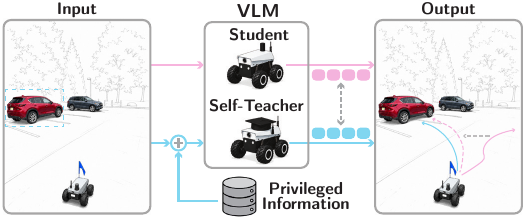}
    \vskip -0.06in
    \caption{\textbf{Action self-distillation.} The self-teacher uses privileged ground-truth trajectory feedback to extract scene semantics, infer a language goal, and generate reasoning traces explaining the demonstrated behavior. Blue arrows denote the feedback-conditioned information flow to the self-teacher, while pink arrows denote the student pathway. The student observes only deployable inputs and learns to align its reasoning and trajectory tokens with the self-teacher, removing the need for external goal or reasoning annotations.}
    \label{fig:method_cartoon}
    \vspace{-0.15in}
\end{figure}

Prior approaches address this gap either by relying on human annotators for high-quality language labels~\citep{hirose2025omnivla, hirose24lelan, chiang2024mobility, castro2025vamos}, or by orchestrating multiple external models to generate goals, semantic descriptions, or reasoning supervision. 
Departing from these approaches, we investigate whether a single pretrained VLM can supervise itself from trajectory-only navigation data. 
This motivates our research question: \textit{How can we learn language goal-conditioned end-to-end navigation policies from trajectory-only robot datasets, without relying on dataset language annotations or multi-model annotation pipelines?}



In human interaction with frontier VLMs, feedback plays a central role in refining model behavior when initial responses do not match the user's intent. A key observation is that when a model is conditioned on corrective or informative feedback about its own outputs, it can recover coherent goal-aligned predictions even when it could not do so from the input alone~\citep{zelikman2022star, zhao2026self, hsieh2023distilling}. Inspired by this feedback-driven reasoning capability, we treat trajectory feedback as privileged information during training. Given such feedback, a VLM can infer the underlying goal, extract relevant scene semantics, and produce planning-oriented reasoning during training. To this end, we propose Training Generalist Navigation Policies via Action Self-Distillation (\algname), a framework in which a single model extracts language goals from trajectory-conditioned scene reasoning as self-teacher, and then learns to generate goal-conditioned trajectories together with corresponding reasoning traces. 




\algname{} instantiates this idea through a self-distillation procedure with two passes. In the teacher pass, the VLM receives the current observation together with ground-truth future trajectory, projected onto an egocentric observation image. This ground truth trajectory as privileged feedback information guides the model to produce structured perception-prediction-planning reasoning and to extract a natural language goal. In the student pass, the model is trained to condition on the extracted goal and predict both future actions and structured reasoning from onboard observations alone. The self-teacher is implemented as an EMA-updated copy of the student~\cite{grill2020bootstrap}, allowing language-goal and reasoning supervision to be generated within the training loop. Critically, \algname{} operates directly on raw image-trajectory pairs, requiring no human-provided language goals or reasoning annotations. To our knowledge, this is the first application of feedback-conditioned self-distillation on robot control, converting existing unannotated navigation datasets into goal-conditioned reasoning supervision without external labeling.

We deploy \algname{} on a FrodoBot rover in zero-shot manner across a range of out-of-distribution natural language goals in both indoor and outdoor environments. The policy is trained on a mixture of RECON~\cite{shah2021rapid} and FrodoBot-2K~\citep{frodobots2k}, without any language goal or reasoning annotations. Across 9 indoor and 4 outdoor language-conditioned real-world experiments, we compare \algname{} against competitive VLA navigation baselines, including OmniVLA~\cite{hirose2025omnivla} and NaVILA~\citep{cheng2024navila}. \algname{} demonstrates language goal steerability across directional, color-based, and object-centric commands, and further generalizes to both multi-stage and indoor navigation tasks that are \emph{unseen} during training. Across all 13 real-world tasks, \algname{} achieves the highest overall success rate, outperforming the strongest baseline, OmniVLA, by $30\%$ in the indoor setting and reaching a $95\%$ success rate in the outdoor setting. On multi-stage tasks, \algname{} further improves over the leading baseline, NaVILA, by $30\%$.

\section{Related Work}\label{sec:related}

\textbf{Learning-based Ground Navigation.}
Early learning-based navigation networks like BADGR~\citep{kahn2021badgr} showed that ground robots can learn navigation directly from real-world experience, moving away from hand-designed cost functions.
ViNG~\citep{shah2021ving} and RECON~\citep{shah2021rapid} further demonstrated visual-goal navigation and open-world exploration by combining local navigation with topological information or latent goals.
More recent foundation-style navigation policies, including GNM~\citep{shah2023gnm}, ViNT~\citep{shah2023vint}, and NoMaD~\citep{sridhar2024nomad}, scale this idea by training goal-conditioned visual navigation policies on heterogeneous robot data, with NoMaD introducing diffusion policies and goal masking to unify goal-directed navigation and exploration.
While these methods demonstrate strong large-scale learned navigation capabilities, they remain limited to visual goal conditioning and do not support natural language instructions, which are easier to specify and more practical for real-world deployment.

\textbf{Natural Language Goals and Reasoning in Navigation.}
Language has long been studied as a natural interface for robot instruction following and vision-and-language navigation (VLN), where agents map free-form commands to navigation actions from visually grounded observations \citep{anderson2018vision,gu2022vision,ku2020room,lin2025evolvenav}. Recent work has explored language as a goal modality for robot navigation, recognizing that natural language is the most accessible interface for human control. LM-Nav~\citep{shah2023lm} composes an LLM, CLIP-style image-language grounding, and a visual navigation model to follow language instructions without language-annotated robot data. NavGPT~\citep{zhou2024navgpt} and NavGPT-2~\citep{zhou2024navgpt2} study explicit LLM reasoning for vision-and-language navigation, while NaVILA~\citep{cheng2024navila} and VAMOS~\citep{castro2025vamos} extend this toward real robot navigation with embodiment-specific planning modules. More recently, OmniVLA~\citep{hirose2025omnivla} and LeLaN~\citep{hirose24lelan} achieve language-conditioned navigation on real robots by relying on external labeling pipelines to supply language annotations at training time. While these works demonstrate that language conditioning can benefit navigation, none address the annotation bottleneck from within the training loop. Our approach instead elicits language goals and contextual descriptions entirely through self-distillation, requiring no external annotation of any kind.

\textbf{Reasoning Distillation and Self-Distillation.}
Our framework builds on EMA-based self-supervised learning~\citep{grill2020bootstrap}, which showed that models can supervise themselves under richer context without external labels.
STaR~\citep{zelikman2022star} extended this idea to language by bootstrapping reasoning from self-generated rationales, while related methods~\citep{zhao2026self, hubotter2026reinforcement, shenfeld2026self} demonstrated that distilling a privileged-context teacher into a deployment-time student is effective for capability acquisition.
The closest control-domain analogue is SelfD~\citep{zhang2022selfd}, which pseudo-labels driving videos with action supervision and retrains a student policy, but predicts actions rather than language goals or contextual descriptions.
Recent embodied reasoning works~\citep{feng2025verdi, xu2024vlm, chen2020learning} similarly use privileged VLM reasoning to improve deployable policies, but rely on separate external teacher models.
In contrast, we use the same open-source VLM as both teacher and student, grounding it through annotated trajectory feedback to elicit language goals and contextual descriptions within the training loop. To our knowledge, this is the first application of self-distillation as a grounding mechanism for robot control.

\vspace{-0.05in}

\section{Training Generalist Navigation Policies via Action Self-Distillation}\label{sec:method}
\vspace{-0.05in}
We propose a novel method for training robotic navigation policies via Action Self-Distillation. Policies trained with \algname{} can navigate to goals specified by natural language, while also reasoning about their actions. We briefly introduce notation and formulation, and then describe \algname{}.

\begin{figure}[t]
\vspace{-0.4in}
    \centering
    \includegraphics[width=0.9\linewidth]{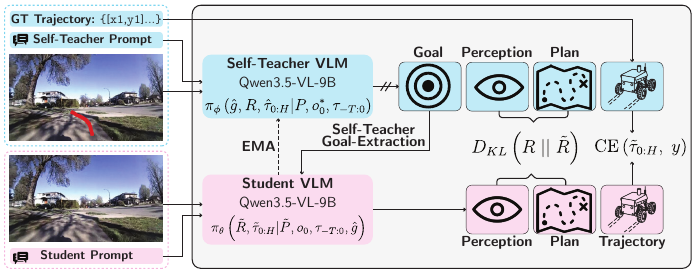}
    \vskip -0.06in
    \caption{\textbf{Action Self-Distillation.} A single pre-trained VLM is used as both the self-teacher $\pi_{\phi}$ and student $\pi_{\theta}$. During training, the self-teacher receives privileged information by projecting the ground-truth future trajectory $y$ onto the front-view observation as $o^{*}_{0}$, enabling it to infer a language goal $\hat{g}$ and scene-grounded reasoning trace $R$. The student observes only deployable inputs $o_0$ and learns to match the teacher's reasoning tokens with a KL loss, while predicting future trajectory tokens $\tilde{\tau}_{0:H}$ with a cross-entropy loss against the ground-truth trajectory $y$. The teacher weights are updated as an EMA of the student.
    }
    \label{fig:method_cartoon}
    \vspace{-8pt}
\end{figure}

\subsection{Problem Formulation}
We consider a ground robot that navigates using ego-centric single-view observations.
At time $t$, the robot receives a front-camera image $o_t$, and its state as a two-dimensional birds-eye view location $s_t = [x, y]\in\mathbb{R}^2$. 
Given a history of $T$ past states $\tau_{-T:0} = \{s_{-T}, \ldots, s_0\}$, the goal of our policy is to predict a future trajectory $\tau_{0:H} = \{s_0, \ldots, s_H\}$ over a planning horizon $H$. We build on a pretrained vision-language model (VLM) $\pi$ that autoregressively produces text tokens; trajectories are serialized in a JSON-style text format and tokenized by the language model, enabling trajectory prediction via next-token prediction.
We seek a policy that additionally emits a \emph{structured reasoning trace}
$R$ and can be conditioned on a natural-language goal $g$:
\begin{equation}
    \hat{\tau}_{0:H},\; R \;\sim\; \pi\!\left(\,\cdot \mid o_{-T:0},\,
    \tau_{-T:0},\, g\right).
    \label{eq:student_policy}
\end{equation}
Crucially, neither reasoning annotations nor goal labels exist in any of
the navigation datasets we train on~\cite{shah2021rapid,frodobots2k,hirose24lelan,shah2023gnm};
all language supervision is generated by the method itself.


\subsection{Training: Language Following and Reasoning via Action Self-Distillation}



We use the ground-truth trajectory as a privileged \textit{feedback information} to elicit structured semantics and goal information.
This privileged feedback conditioned forward pass constitutes the \emph{self-teacher}, which generates both a structured reasoning trace and a semantic navigation goal.
The \emph{student} is trained using only robot observations, conditioned on the language goal extracted by the self-teacher, and supervised to generate both the reasoning trace and the future trajectory. At inference time, the privileged trajectory input is removed and the extracted training-time goal is replaced by a free-form language goal provided by a human operator, enabling language-conditioned navigation without requiring human-annotated reasoning or goal labels during training.
Both teacher and student are instances of the same pretrained VLM; the self-teacher weights $\phi$ are maintained as an exponential moving average (EMA) of the student weights $\theta$: $\phi \leftarrow \alpha\,\phi + (1-\alpha)\,\theta$ .

\textbf{Self-Teacher Pass.}
As listed in Line \algline{line:teacher_pass} of Algorithm~\ref{alg:asd-training}, the self-teacher $\pi_\phi$ receives the current camera view $o_0$ augmented with a polyline overlay of the ground-truth trajectory $y = \tau_{0:H}$ (annotated observation $o^{*}_{0}$), and produces language goal $\hat{g}$, reasoning $R$, and trajectory $\hat{\tau}_{0:H}$ with prompt $P$ as
\begin{equation}
    \hat{g},\; R,\; \hat{\tau}_{0:H}
    \;\sim\;
    \mathrm{sg}\!\left(\pi_\phi\!\left(\,\cdot \mid P,\, o^{*}_{0}, \, \tau_{-T:0}\right)\right).
    \label{eq:teacher}
\end{equation}
 
\textbf{Student Pass.}
The student pass is summarized in Line \algline{line:student_pass} of Algorithm~\ref{alg:asd-training}. Specifically, the student $\pi_\theta$ receives the image observation $o_0$, state history $\tau_{-T:0}$, and the self-teacher's extracted goal $\hat{g}$, and produces a reasoning trace $\tilde{R}$ and trajectory with the mirrored prompt $\tilde{P}$ as
\begin{equation}
    \tilde{R},\; \tilde{\tau}_{0:H}
    \;\sim\;
    \pi_\theta\!\left(\,\cdot \mid \tilde{P},\, o_0,\, \tau_{-T:0},\, \hat{g}\right).
    \label{eq:student}
\end{equation}

\textbf{Training Objective.}
The output token sequence is partitioned at the onset of the trajectory block.
Tokens in the reasoning span (Perception, Agents, Prediction, Planning) are supervised by a KL divergence loss against the self-teacher's logits, that is
\begin{equation}
    \mathcal{L}_{\mathrm{reason}}
    = D_{\mathrm{KL}}\!\left(\, R \;\|\; \tilde{R}\,\right).
    \label{eq:reason_loss}
\end{equation}
Tokens in the trajectory span are supervised by cross-entropy against the ground-truth trajectory
\begin{equation}
    \mathcal{L}_{\mathrm{traj}}
    = \mathrm{CE}\!\left(\tilde{\tau}_{0:H},\; y\right).
    \label{eq:ce_loss}
\end{equation}
The combined objective is $\mathcal{L} = \mathcal{L}_{\mathrm{traj}} + \mathcal{L}_{\mathrm{reason}}$, summarized in Line \algline{line:loss_reason}-\algline{line:loss_total} of Algorithm \ref{alg:asd-training} .


\textbf{Structured Reasoning Format.}
The self-teacher prompt enforces a fixed output structure: \textit{Task} (navigation goal; goal-conditioned mode only), \textit{Perception} (visible terrain and structures), \textit{Agents} (moving entities present in the scene, or ``none''), \textit{Prediction} (expected agent motion), and \textit{Planning} (a discrete action from a fixed vocabulary, justified by named scene features), followed by the executable trajectory.
The structured format constrains the teacher to ground its reasoning in observable scene content; the teacher prompt explicitly instructs the model to describe only what is visible in the raw image, so that any claim can be verified against the unannotated observation.
Full prompt templates for each dataset are provided in the Section~\ref{appendix: training_detail} of supplementary material.

\subsection{Deployment: Navigating to Language Goals}

At deployment, the self-teacher is absent and no ground-truth trajectory is available. The student policy $\pi_\theta^\star$ plans a trajectory and reasoning trace directly from the current observation and history as
\begin{equation}
    \tilde{R},\; \tilde{\tau}_{0:H}
    \;\sim\;
    \pi_\theta^\star\!\left(\,\cdot \mid \tilde{P},\, o_0,\, \tau_{-T:0}\right).
    \label{eq:inference}
\end{equation}
In the goal-conditioned setting, a human operator supplies a natural-language goal $g^\circledast$ in place of the self-teacher's extracted $\hat{g}$ via
\begin{equation}
    \tilde{R},\; \tilde{\tau}_{0:H}
    \;\sim\;
    \pi_\theta^\star\!\left(\,\cdot \mid \tilde{P},\, o_0,\, \tau_{-T:0},\, g^\circledast\right).
    \label{eq:inference_gc}
\end{equation}
The goal $g^\circledast$ may take several forms, including a path description (e.g.\ \texttt{``navigate along the sidewalk''}), or an object reference (e.g.\ \texttt{``navigate towards the red building''}).
Because the student is trained on goals $\hat{g}$ produced by the self-teacher from diverse scenes across the full training distribution, this constitutes zero-shot language-conditioned navigation.
The model generalizes to operator-supplied instructions at test time without ever seeing human-specified goals during training.

\vspace{-0.1in}
\begin{algorithm}   
\caption{Training Policies via Action Self-Distillation}
\label{alg:asd-training}

\KwIn{Trajectory-only dataset $\mathcal{D}$, student $\pi_\theta$, self-teacher $\pi_\phi$, EMA rate $\alpha$}
\KwOut{Trained goal-conditioned navigation policy $\pi_\theta$}

Initialize $\theta$ from a pretrained VLM and set $\phi \leftarrow \theta$\;

\For{each training iteration}{
    Sample $(o_0, \tau_{-T:0}, y)$ from $\mathcal{D}$, where $y=\tau_{0:H}$\;
    Project $y$ onto $o_0$ to obtain privileged observation $\tilde{o}_0$\;

    $(\hat{g}, R, \hat{\tau}_{0:H}) \sim \pi_\phi(\cdot \mid P, o^*_0, \tau_{-T:0})$, Eq.~\eqref{eq:teacher}\nllabel{line:teacher_pass}\;
    $(\tilde{R}, \tilde{\tau}_{0:H}) \sim \pi_\theta(\cdot \mid \tilde{P}, o_0, \tau_{-T:0}, \hat{g})$, Eq.~\eqref{eq:student} \nllabel{line:student_pass}\;

    $\mathcal{L}_{\mathrm{reason}} \leftarrow D_{\mathrm{KL}}(R \,\|\, \tilde{R})$, Eq.\eqref{eq:reason_loss} \nllabel{line:loss_reason}\;
    $\mathcal{L}_{\mathrm{traj}} \leftarrow \mathrm{CE}(\tilde{\tau}_{0:H}, y)$, Eq.\eqref{eq:ce_loss}\;
    $\mathcal{L} \leftarrow \mathcal{L}_{\mathrm{traj}} + \mathcal{L}_{\mathrm{reason}}$ \label{line:loss_total}\;

    Update $\theta$ by minimizing $\mathcal{L}$\;
    $\phi \leftarrow \alpha \phi + (1-\alpha)\theta$\;
}
\end{algorithm}
\vspace{-0.1in}

\section{Experiments}\label{sec:experiment}
\begin{table*}[b]
\vspace{-0.1in}
\centering
\caption{\textbf{Indoor Language-Conditioned Navigation.} We evaluate tasks across three categories: directional, object-centric, and multi-stage navigation. Difficulty level increases with the number of possible directions for directional tasks, and the number of objects in the scene for object-centric tasks. In the multi-stage category, \texttt{Doorway} requires the robot to pass through a door, while \texttt{Hallway} further requires it to turn right after the doorway and move along the hallway. We use $-$, $+$, and $*$ to denote easy, medium, and hard tasks. More detailed task descriptions and visualizations are provided in Section~\ref{appendix:indoor_task} of the supplementary material. 
}
\label{tab:indoor_tasks}
\footnotesize
\renewcommand{\arraystretch}{1.12}
\setlength{\tabcolsep}{1.5pt}
\vskip -0.06in
\resizebox{\textwidth}{!}{
\begin{tabular}{lccccccccc c}
\toprule
\multirow{3}{*}{\textbf{Method}}
&
\multicolumn{3}{c}{\textbf{Directions}}
&
\multicolumn{4}{c}{\textbf{Objects}}
&
\multicolumn{2}{c}{\textbf{Multi-stage}}
&
\multirow{3}{*}{\textbf{Avg.}}
\\
\cmidrule(lr){2-4}
\cmidrule(lr){5-8}
\cmidrule(lr){9-10}
&
\makecell{Straight Door$^-$\\(1 direction)}
&
\makecell{Left chair$^+$\\(2 directions)}
&
\makecell{Right chair$^*$\\(3 directions)}
&
\makecell{Blue$^-$\\chair}
&
\makecell{Trash$^-$\\bin}
&
\makecell{Chair$^+$\\(2 objects)}
&
\makecell{Monitor$^*$\\(3 objects)}
&
Doorway$^+$
&
Hallway$^*$
&
\\
\midrule

OmniVLA
& 0.8 & 0.9 & 0.2
& 0.7 & \textbf{1.0} & 0.4 & 0.2
& 0.7 & 0.0
& 0.54 \\

NaVILA
& 0.3 & 0.2 & 0.1
& 0.4 & 0.1 & 0.3 & 0.2
& 0.3 & 0.5
& 0.27 \\


\midrule

\textbf{Qwen2.5-VL-\algname{}}
& \underline{\smash{0.9}} & \underline{\smash{1.0}} & \underline{\smash{0.8}}
& \underline{\smash{1.0}} & \underline{\smash{0.9}} & \underline{\smash{0.9}} & \underline{\smash{0.7}}
& \underline{\smash{0.8}} & \underline{\smash{0.6}}
& \underline{\smash{0.84}} \\

\textbf{Qwen3.5-VL-\algname{}}
& \textbf{1.0} & \textbf{1.0} & \textbf{0.9}
& \textbf{1.0} & \textbf{1.0} & \textbf{1.0} & \textbf{0.9}
& \textbf{0.9} & \textbf{0.8}
& \textbf{0.94} \\

\bottomrule
\end{tabular}
}
\end{table*}

In this section, we conduct experiments to assess \algname{} by addressing the following two questions: \textbf{Q1:} How well does \algname{} generalize across real-world environments compared with state-of-the-art navigation VLAs? \textbf{Q2:} What capabilities are induced by \algname{}?


\subsection{Experimental Setup}
\algname uses Qwen2.5-VL-3B-Instruct~\cite{yang2024qwen2} as the base VLM, hereafter referred to as Qwen2.5-VL, and is trained through the self-distillation pipeline introduced in Section~\ref{sec:method}, on a mixture of the RECON~\cite{shah2021rapid} and FrodoBot2k~\cite{frodobots2k} datasets. The training details are summarized in Section~\ref{appendix: training_detail} of the Supplementary Material. We evaluate it in real-world settings across both indoor and outdoor environments. We focus on language-conditioned navigation tasks, where user prompts ask the robot to navigate to a specific goal. All experiments are deployed on a FrodoBot EarthRover Mini equipped with a front-facing camera. At each planning step, \algname predicts receding-horizon trajectory, and converted by the robot's built-in API into linear and angular velocity commands for execution.

\subsection{Indoor Language-Conditioned Navigation}\label{sec:indoor}
We conduct indoor real-world navigation experiments in a laboratory environment with typical office objects, including chairs, trash bins, boxes, and monitors. We evaluate 9 tasks covering different levels of difficulty, from simple directional and object-identification goals, such as \texttt{``Navigate to the blue chair''}, to multi-stage goals, such as \texttt{``Navigate through the door, turn right, then follow the hallway''}. For each task, we use fixed language prompts and test each method under four different initializations, where the initial position and orientation of the robot and objects are varied. We compare \algname against state-of-the-art language-conditioned VLAs, OmniVLA~\citep{hirose2025omnivla} and NaVILA~\cite{cheng2024navila}. See Section~\ref{appendix:real_world_task} of the Supplementary for detailed tasks information and evaluation.

Indoor language-conditioned navigation results are reported in Table~\ref{tab:indoor_tasks}. Our \algname achieves consistently high success rates across all tasks and obtains the best average success rate as $84\%$. Among three hard tasks, including \texttt{``Navigate to the right chair''}, \texttt{``Navigate to the monitor''} and \texttt{``Navigate through the door, turn right, then follow the hallway''}, \algname{} significantly outperforms leading baseline by $60\%$, $50\%$ and $10\%$ respectively, demonstrating strong generalization capability in complex tasks.

The task performance of \algname{} remains relatively consistent compared with competitive baselines. Across all evaluated tasks, \algname{} achieves success rates above $50\%$. In contrast, OmniVLA and NaVILA obtain substantially lower success rates on several tasks and even exhibit near-total failure in certain scenarios, such as OmniVLA, which achieves $0\%$ success in \texttt{Hallway}. These results suggest that the baselines have limited robustness to specific navigation scenarios, which we further analyze in Section~\ref{sec:assessment}.


\begin{wraptable}{R}{0.5\textwidth}
\vspace{-12pt}
\centering
\caption{
\textbf{Outdoor Language-Conditioned Navigation.} We evaluate on 4 different tasks, including \texttt{navigate along the sidewalk, navigate to the statue/tree/car}. More detailed task descriptions and visualizations are provided in Section~\ref{appendix:outdoor_task} of the supplementary material. }
\label{tab:outdoor_tasks}
\footnotesize
\renewcommand{\arraystretch}{1.12}
\setlength{\tabcolsep}{3.5pt}
\vspace{-0.06in}
\resizebox{0.5\textwidth}{!}{
\begin{tabular}{lccccc}
\toprule
\textbf{Method}&Sidewalk&Statue&Tree&Car&Avg.
\\
\midrule
OmniVLA
& \textbf{1.0} & 0.6 & \textbf{1.0} & \textbf{1.0} & 0.90\\
NaVILA
& 0.9 & 0.1 & 0.0 & 0.3 & 0.33\\
\midrule
\textbf{Qwen2.5-VL-\algname{}}
& \textbf{1.0} & 0.8 & \textbf{1.0} & \textbf{1.0} & 0.95\\
\textbf{Qwen3.5-VL-\algname{}}
& \textbf{1.0} & \textbf{0.9} & \textbf{1.0} & \textbf{1.0} & \textbf{0.98}\\
\bottomrule
\end{tabular}
}
\vspace{-12pt}
\end{wraptable}


\subsection{Outdoor Language-Conditioned Navigation}
We conduct outdoor real-world navigation experiments in a campus environment. We evaluate 4 tasks, including \texttt{navigate along the sidewalk}, and \texttt{navigate to the statue, tree and car}. We follow the same experimental setup and compare to the same baselines as in the indoor setting. See Supplementary Section~\ref{appendix:real_world_task} for tasks and evaluation details. Results are reported in Table~\ref{tab:outdoor_tasks}, where we observe that \algname achieves consistently high success rates across all tasks and obtains the best average success rate of $95\%$. In contrast, the baselines either show lower success rates on specific tasks, such as OmniVLA achieving 60$\%$ on \texttt{Statue}, or frequently fail on several tasks, such as NaVILA achieving 0$\%$ on \texttt{Tree} and 10$\%$ on \texttt{Statue}. We further discuss the result in Section~\ref{sec:assessment}.



\subsection{Behavior Assessment}\label{sec:assessment}
In this section, we provide a detailed assessment and discussion of \algname's learned capabilities, qualitative behavior and relative performance gains compared to baselines.

\textbf{Zero-Shot Navigation to Out-of-Distribution Objects and Goals.} Indoor navigation as described in Sec.~\ref{sec:indoor} is evaluated in Table~\ref{tab:indoor_tasks}, highlighting \algname{}'s ability to generalize to out-of-distribution (OOD) objects, goals, and environments. In particular, our indoor experiments involve laboratory objects and indoor navigation goals that are absent from the training data, since both RECON and FrodoBot-2K are primarily outdoor navigation datasets. Despite this large domain shift, \algname{} maintains 84$\%$ average success, indicating strong generalization to novel object categories, language goals, and indoor environments. We observe a similar trend in the outdoor experiments in Table~\ref{tab:outdoor_tasks}: \texttt{"navigate to the statue"} requires grounding an OOD object category, yet \algname{} achieves an $80\%$ success, outperforming OmniVLA by $20\%$ and NaVILA by $70\%$.

\begin{figure}[t]
\vspace{-0.4in}
    \centering
    \includegraphics[width=0.95\linewidth]{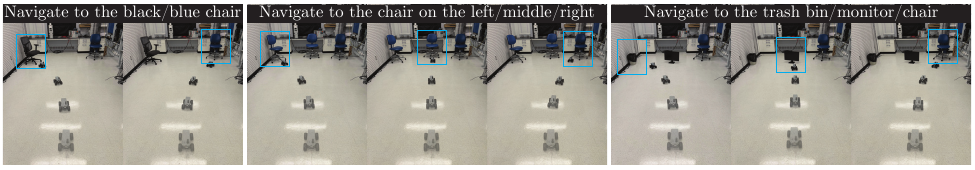}
    \caption{\textbf{Steerability Demonstration}. We demonstrate \algname's steerability under color (left), direction (middle), and object (right) prompts. Within the same setting, our method can be directed to different targets.}
    \label{fig:steerability}
    \vspace{-0.15in}
\end{figure}

\textbf{Steerability.}
\algname{} is steerable with different language goals within the same environment. Fig.~\ref{fig:steerability} shows that \algname{} can be directed by different \texttt{Directions}, interpreting directional commands, and different \texttt{Objects}, requiring to ground distinct object categories. In the \texttt{Car} example, \algname{} distinguishes object \textit{attributes}, such as color. These examples validate the generalization and language-following capabilities. This is notable because our training data follows a one-on-one scene and goal supervision: each observation is paired with a single language goal and a single demonstrated trajectory, without counterfactual trajectory augmentation. In contrast, methods such as OmniVLA rely on broader multimodal training and over 9,500 hours of data collected from 10 platforms. Despite using substantially less data, \algname{} generalizes to produce distinct trajectories conditioned on different language goals in the same scene, demonstrating more data-efficient steerable behavior.


\begin{figure}[t]
    \centering
    \includegraphics[width=0.95\linewidth]{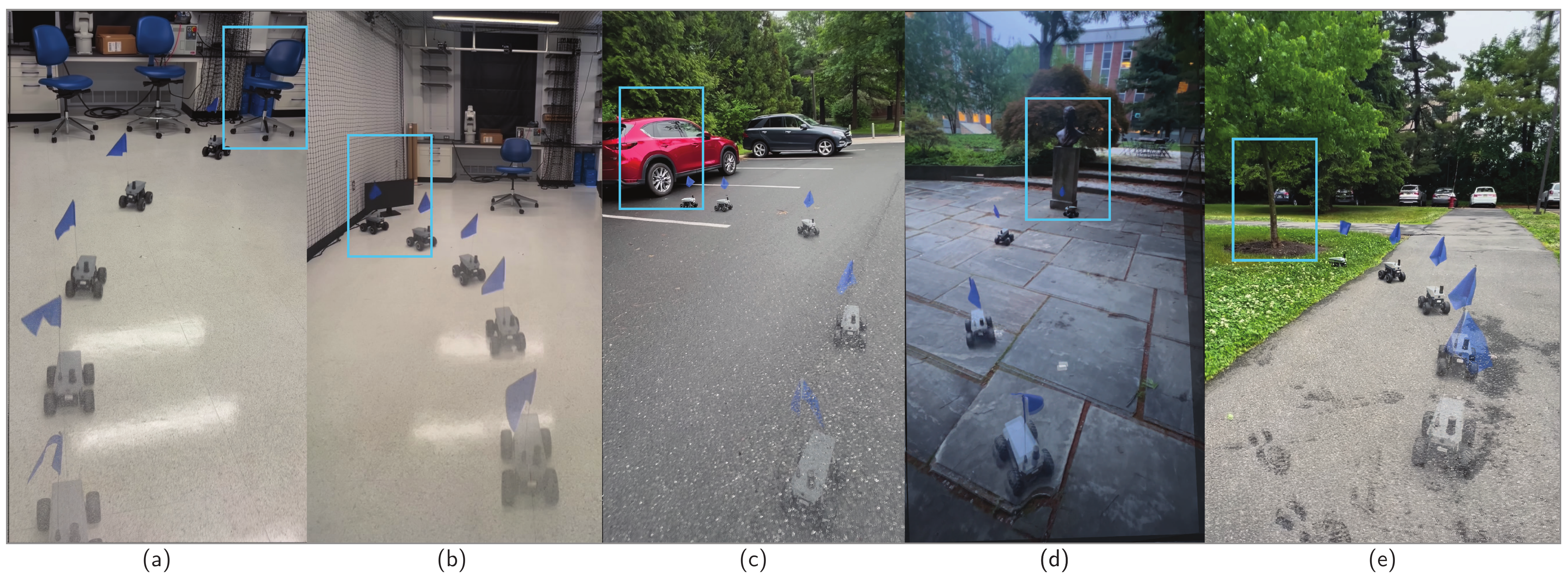}
    \vspace{-0.6em}
    \caption{\textbf{Single-Stage Navigation Across Diverse Indoor and Outdoor Environments.} We demonstrate \algname{}'s capability to execute single-stage navigation goals specified in natural language. The examples are ordered from left to right as two indoor tasks (\texttt{right chair} and \texttt{monitor}) and three outdoor tasks (\texttt{car}, \texttt{statue}, and \texttt{tree}). Full videos are provided in the Supplementary Material.}
    \label{fig:outdoor}
    \vspace{-0.25in}
\end{figure}

\begin{figure}[b]
    \vspace{-0.2in}
    \centering
    \includegraphics[width=\linewidth]{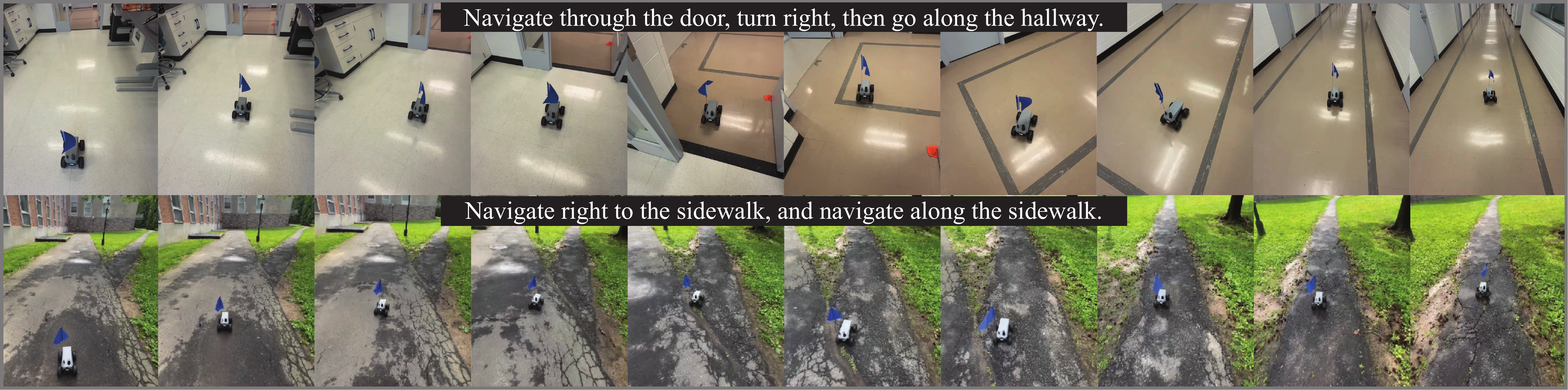}
    \caption{\textbf{Multi-Stage Reasoning Capabilities for Outdoor and Indoor Navigation Tasks.} We demonstrate \algname's capability to execute complex, multi-stage navigation goals. The first row contains an indoor navigation demonstration, while the bottom row contains an outdoor navigation demonstration, both with their respective tasks described on the figure. Full videos are provided in the Supplementary Material.}
    \label{fig:multistep}
\end{figure}

\textbf{Multi-Stage Task Execution Capabilities.}
\algname is capable of executing complex, multi-stage goals, as evaluated in the \texttt{Multi-stage} tasks in Table~\ref{tab:indoor_tasks}. \texttt{Doorway} is a two-stage task where the robot needs to navigate to the door and go through it. \texttt{Hallway} is a four-stage task where the robot needs to further turn right after passing the door and move along the hallway. Our \algname achieves success rates of 80$\%$ and 60$\%$ on the shorter and longer tasks, respectively. In contrast, OmniVLA fails on the longer task with a 0$\%$ success rate, while NaVILA performs poorly on the shorter task with a 30$\%$ success rate. Figure~\ref{fig:multistep} reports additional outdoor and indoor multi-stage navigation tasks. 

\begin{figure}[t]
\vspace{-0.2in}
    \centering
    \includegraphics[width=\linewidth]{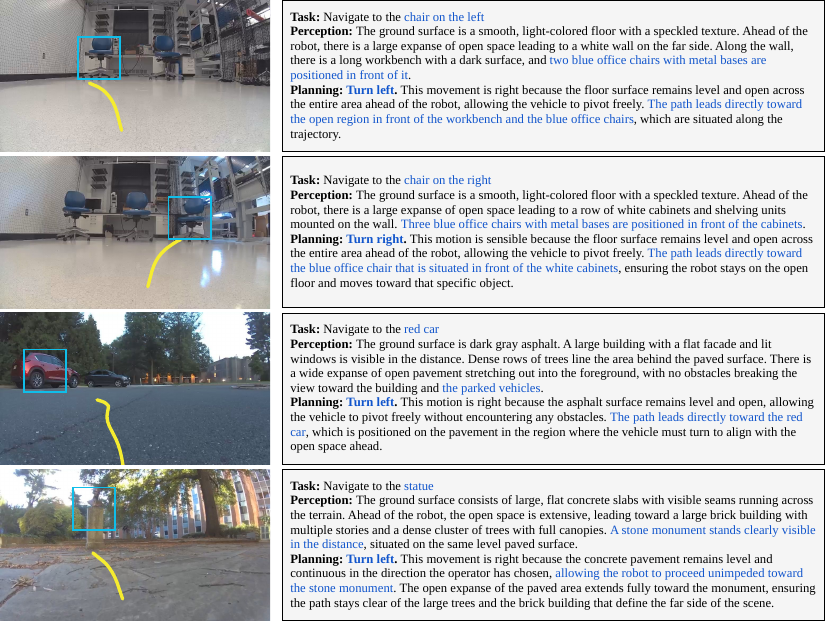}
    \vspace{-5mm}
    \caption{\textbf{Reasoning Legibility.} We present representative findings from our real-world experiments, showing predicted trajectories alongside the corresponding highlighted reasoning traces. These examples demonstrate that the generated reasoning is interpretable and well aligned with the predicted robot behavior.}
    \label{fig:reasoning_example}
    \vspace{-0.25in}
\end{figure}



\textbf{Reasoning.}
Figure~\ref{fig:reasoning_example} shows that \algname{}'s reasoning aligns with both the goal and the predicted trajectory. In the indoor examples, the model reasons to turn left or right toward the requested chair. In the outdoor examples, it reasons to turn left toward the red car or the statue, relating each planned motion to the target location and the open space ahead.



\begin{wraptable}{R}{0.38\textwidth}
\vspace{-0.17in}
\centering
\caption{
\textbf{Task success rates and completion times of foundation models on indoor navigation task}. We extend the evaluation of \texttt{Left chair} in Table~\ref{tab:indoor_tasks} by including Qwen2.5-VL and Claude Opus 4.7, and report the task completion time for all methods. We omit the completion time for Qwen2.5-VL because its predicted actions are generally physically infeasible.}
\label{tab:foundation_model}
\vspace{-5pt}
\footnotesize
\renewcommand{\arraystretch}{1.12}
\setlength{\tabcolsep}{3.5pt}
\begin{tabular}{lcc}
\toprule
\textbf{Method}& SR $\uparrow$ & Time (s) $\downarrow$\\
\midrule
OmniVLA & 0.9 & 25.3\\
NaVILA & 0.2 & 28.9\\
\midrule
Qwen2.5-VL
& 0.0 & --\\
Claude Opus 4.7
& 0.7 & 846.1 \\
\midrule
\textbf{Qwen2.5-VL-\algname}
& \textbf{1.0} &  \textbf{15.3} \\
\bottomrule
\end{tabular}
\vspace{-0.22in}
\end{wraptable}

\textbf{Foundation Model Evaluations.} We perform zero-shot evaluation with vanilla Qwen2.5-VL and Claude Opus 4.7 on the \texttt{Left chair} task. Qwen2.5-VL yields $0\%$ success rates, producing degenerate trajectories. 
Claude Opus 4.7 generates plausible reasoning and correctly identifies navigation targets, achieving $70\%$ success rate. However, its inference latency reaches several minutes per planning step, leading to an extremely slow average completion time of 846.1 seconds. This renders it impractical for real-world navigation within a reasonable time frame. In contrast, \algname achieves $100\%$ success rate while maintaining a relatively fast completion time. These observations suggest that frontier VLMs possess strong semantic reasoning, but still require a grounding to translate this into actionable, low-latency robot behavior. Our approach provides this grounding mechanism.

\textbf{Importance of Reasoning.} We ablate the reasoning traces in Table~\ref{tab:frozen_teacher}. The performance of Qwen2.5-VL-\algname (SR 80\%) decreases when we disable the reasoning traces at training (SR 26\%) or inference (SR 14\%), demonstrating the importance of reasoning.

\textbf{Frozen-teacher Baselines \& Benefit of EMA Self-Distillation.} We ablate the teacher in self-distillation by comparing against offline-label SFT baselines trained with annotations from Claude Opus 5 or Qwen2.5-VL, as shown in Table~\ref{tab:frozen_teacher}. We find that Qwen2.5-VL-\algname (SR 80\%) outperforms offline labeling with Qwen2.5-VL (SR 22\%) and Opus 5 (SR 36\%). Using the same training dataset (RECON + FrodoBots-2K) as \algname, we fine-tune NaVILA, which achieves only a 16\% success rate. These empirically support the benefit of EMA self-distillation over SFT on teacher pre-annotated pseudo-labels, consistent with findings that teacher-student capacity mismatch degrades distillation performance~\citep{li2025small, zhang2025towards}. Additionally, Qwen3.5-VL-\algname (SR 92\%) achieves better performance with a stronger backbone.

\begin{table}[t]
\centering
\caption{\textbf{Frozen-teacher baselines and reasoning ablations.} Success rate on representative tasks from Table~\ref{tab:indoor_tasks} and Table~\ref{tab:outdoor_tasks} requiring directional, object-level, multi-stage, and outdoor reasoning. \emph{Goal-only supervision} trains on self-teacher-inferred goals without reasoning traces. \emph{Reasoning off at inference} disables reasoning at test time. \emph{Labels + SFT} baselines fine-tune on offline annotations from a frozen Claude Opus~5 or Qwen2.5-VL teacher. Best results are in \textbf{bold}, second best are \underline{underlined}.}
\label{tab:frozen_teacher}
\small
\setlength{\tabcolsep}{6pt}
\vspace{2pt}
\resizebox{\textwidth}{!}{
\begin{tabular}{lcccccc}
\toprule
\multirow{2}{*}{\textbf{Method}} & \textbf{Directions} & \multicolumn{2}{c}{\textbf{Objects}} & \textbf{Multi-stage} & \textbf{Outdoor} & \multirow{2}{*}{\textbf{Avg.}} \\
\cmidrule(lr){2-2}\cmidrule(lr){3-4}\cmidrule(lr){5-5}\cmidrule(lr){6-6}
 & Right chair & Chair & Monitor & Hallway & Car & \\
\midrule
Qwen2.5-VL-\algname{} {\scriptsize (goal-only supervision)} & 0.4 & 0.2 & 0.2 & 0.1 & 0.4 & 0.26 \\
Qwen2.5-VL-\algname{} {\scriptsize (reasoning off at inference)} & 0.1 & 0.2 & 0.2 & 0.0 & 0.2 & 0.14 \\
\midrule
Claude Opus~5 labels + SFT & 0.3 & 0.5 & 0.3 & 0.3 & 0.4 & 0.36 \\
Qwen2.5-VL labels + SFT & 0.4 & 0.3 & 0.3 & 0.0 & 0.1 & 0.22 \\
NaVILA {\scriptsize (RECON + FrodoBots-2K SFT)} & 0.0 & 0.1 & 0.2 & 0.3 & 0.2 & 0.16 \\
\midrule
Qwen2.5-VL-\algname{} & \underline{0.8} & \underline{0.9} & \underline{0.7} & \underline{0.6} & \textbf{1.0} & \underline{0.80} \\
Qwen3.5-VL-\algname{} & \textbf{0.9} & \textbf{1.0} & \textbf{0.9} & \textbf{0.8} & \textbf{1.0} & \textbf{0.92} \\
\bottomrule
\end{tabular}
}
\vspace{-0.15in}
\end{table}
\section{Conclusion}
We present Training Generalist \textbf{N}avigation Policies via \textbf{A}ction \textbf{S}elf-\textbf{D}istillation, a method for learning instruction-following navigation policies from trajectory-only robot data. Using privileged trajectory feedback during training, \algname{} generates language goals and structured reasoning without human annotations, enabling the student policy to follow complex natural-language instructions and act in the real world. Across $13$ real-world zero-shot tasks on a mobile robot, \algname{} demonstrates strong generalization to OOD directions, colors, objects, indoor laboratory environments, and multi-stage tasks. In indoor settings, \algname{} outperforms the strongest baseline, OmniVLA, by $30\%$; in outdoor settings, it achieves $100\%$ success on in-distribution tasks and $80\%$ success on the OOD commands, outperforming OmniVLA by $20\%$ and NaVILA by $70\%$. For multi-stage tasks, \algname{} further improves over the leading baseline, NaVILA, by $30\%$, despite being trained on significantly less data. Akin to training reasoning-enabled foundation models of language, we hope that \algname serves as a generic recipe for training generalist robot policies with reasoning without requiring expensive human annotations.

\noindent \textbf{Limitations.} While \algname{} removes the need for language annotations, it still relies on demonstration trajectories as privileged feedback; future work could replace these with interactive human feedback for more precise task-level guidance. The quality of self-generated goals and reasoning also depends on the pretrained VLM and may degrade in ambiguous scenes, with small targets, or under fine-grained spatial instructions.


\section*{Acknowledgments}
This work and Felix Heide were supported by an NSF CAREER Award (2047359), Packard and Sloan Fellowships, and research awards from Samsung, Project X, Amazon Science, and Bosch. Felix Heide is a co-founder of Algolux (now Torc Robotics) and Cephia AI, and Head of AI at Torc Robotics.


\newpage
\bibliography{ref}
\newpage
\appendix
\clearpage
\section{Comparing \algname{} vs. Supervised Finetuning}
\phantomsection
\addcontentsline{toc}{section}{S\quad Comparing \algname{} vs. Supervised Finetuning}
\label{app:theory-gen}

In this section, we provide further insight into the experimental findings reported in the main paper. We observe that \algname{}, despite being fine-tuned on substantially less data ($\sim100$ hours demonstration data (\algname{}) v.s. $\sim9500$ hours demonstration data (OmniVLA)) than OmniVLA and NaVILA, exhibits better generalization on indoor out-of-distribution evaluations where \algname{} has not been explicitly fine-tuned. 

In the following, we provide insight into why supervising the reasoning span of our policy with the self-distillation loss $\mathcal{L}_{\mathrm{reason}}$ in Eq.~(4) leads to better generalization and retention of the pretrained VLM's capabilities than standard offline imitation.

\subsection{Privileged Soft-label Distillation Objective}
During training, the self-teacher $\pi_\phi$ generates a reasoning trace using privileged trajectory feedback
\begin{equation}
    R \;\sim\;
    \mathrm{sg}\!\left[
    \pi_\phi\!\left(\cdot \mid P, o^*_0, \tau_{-T:0}\right)
    \right],
    \label{eq:gen-teacher-sample}
\end{equation}
where $o^*_0=f(o_0,y)$ is the annotated observation obtained by projecting the ground-truth future trajectory $y=\tau_{0:H}$ onto the front-camera observation $o_0$. The stop-gradient prevents the reasoning loss from updating the self-teacher directly. The student receives only deployable inputs and the self-teacher's extracted goal $\hat{g}$ as
\begin{equation}
    \tilde{R}
    \;\sim\;
    \pi_\theta\!\left(
    \cdot \mid \tilde{P}, o_0, \tau_{-T:0}, \hat{g}
    \right).
\end{equation}

The reasoning span is supervised by a forward-KL distillation objective

\begin{equation}
    \mathcal{L}_{\mathrm{reason}}
    = D_{\mathrm{KL}}\!\left(\, R \;\|\; \tilde{R}\,\right).
    \label{eq:reason_loss_2}
\end{equation}

\begin{equation}
\mathcal{L}_{\mathrm{reason}}
=
D_{\mathrm{KL}}\!\left(
    \mathrm{sg}\!\left[
    \pi_\phi(\cdot \mid P, o^*_0, \tau_{-T:0})
    \right]
    \,\middle\|\,
    \pi_\theta(\cdot \mid \tilde{P}, o_0, \tau_{-T:0}, \hat{g})
\right),
\label{eq:fwdkl-ours}
\end{equation}
where the KL divergence is applied only over the reasoning-token span.
Equivalently, for the reasoning tokens $R=(r_1,\ldots,r_N)$, this can be written autoregressively as
\begin{equation}
\mathcal{L}_{\mathrm{reason}}
=
\sum_{i=1}^{N}
\mathbb{E}_{R_{<i}\sim
\pi_\phi(\cdot \mid P,o^*_0,\tau_{-T:0})}
\left[
D_{\mathrm{KL}}\!\left(
    \mathrm{sg}\!\left[
    \pi_\phi(\cdot \mid P,o^*_0,\tau_{-T:0},R_{<i})
    \right]
    \,\middle\|\,
    \pi_\theta(\cdot \mid \tilde{P},o_0,\tau_{-T:0},\hat{g},R_{<i})
\right)
\right].
\label{eq:token-fwdkl-ours}
\end{equation}
This form makes explicit that \algname{} transfers the teacher's full token distribution at each reasoning step, rather than only a single hard token.

Expanding Eq.~\eqref{eq:fwdkl-ours} over teacher-generated reasoning traces gives
\begin{equation}
\mathcal{L}_{\mathrm{reason}}
=
\mathbb{E}_{R \sim \pi_\phi(\cdot \mid P,o^*_0,\tau_{-T:0})}
\left[
    \log
    \frac{
    \pi_\phi(R \mid P,o^*_0,\tau_{-T:0})
    }{
    \pi_\theta(R \mid \tilde{P},o_0,\tau_{-T:0},\hat{g})
    }
\right].
\label{eq:expanded-fwdkl-ours}
\end{equation}
Since the teacher distribution is passed into a stop-gradient, the first term is constant with respect to $\theta$.
Thus, optimizing Eq.~\eqref{eq:expanded-fwdkl-ours} is equivalent to minimizing the teacher-weighted cross entropy
\begin{equation}
\mathcal{L}_{\mathrm{reason}}
\equiv
-
\mathbb{E}_{R \sim \pi_\phi(\cdot \mid P,o^*_0,\tau_{-T:0})}
\left[
    \log
    \pi_\theta(R \mid \tilde{P},o_0,\tau_{-T:0},\hat{g})
\right]
+
\mathrm{const}.
\label{eq:teacher-ce-ours}
\end{equation}
The resulting gradient is
\begin{equation}
\nabla_\theta \mathcal{L}_{\mathrm{reason}}
=
-
\mathbb{E}_{R \sim \pi_\phi(\cdot \mid P,o^*_0,\tau_{-T:0})}
\left[
    \nabla_\theta
    \log
    \pi_\theta(R \mid \tilde{P},o_0,\tau_{-T:0},\hat{g})
\right].
\label{eq:fwdkl-grad-ours}
\end{equation}
Therefore, the student is trained to place probability mass on reasoning traces that the self-teacher considers likely.

\subsection{Comparison to Standard SFT}
Standard SFT can also be viewed as matching a behavior-policy distribution.
Let $\pi_b(R \mid \tilde{P},o_0,\tau_{-T:0},\hat{g})$ denote the offline
annotation process that produces reasoning traces. Then SFT minimizes
\begin{equation}
    \mathcal{L}_{\mathrm{SFT}}
    =
    -
    \mathbb{E}_{R^\star \sim \pi_b}
    \left[
        \log
        \pi_\theta(R^\star \mid \tilde{P},o_0,\tau_{-T:0},\hat{g})
    \right]
    =
    H(\pi_b)
    +
    D_{\mathrm{KL}}
    \!\left(
        \pi_b
        \,\middle\|\,
        \pi_\theta
    \right).
\end{equation}
However, in typical offline datasets, $\pi_b$ is observed only through one
hard sampled annotation $R^\star$ per input. Thus, the empirical target
distribution collapses to
\begin{equation}\label{eq:sft}
    \hat{\pi}_b(R \mid \cdot)
    =
    \mathbf{1}[R=R^\star],
\end{equation}
and SFT reduces to hard-label imitation.

In contrast, \algname{} constructs a privileged self-teacher distribution
\begin{equation}
    \pi_\phi(R \mid P,o_0^*,\tau_{-T:0}),
\end{equation}
where $o_0^*=f(o_0,y)$ exposes the demonstrated future trajectory to the teacher. Therefore, the target is not merely an offline annotation sampled from a static behavior policy, but a feedback-conditioned reasoning distribution that explains why the demonstrated action is appropriate for the current observation.

This enables \algname to achieve better generalization, because a given scene may admit multiple valid descriptions and reasonings. For example, in the \texttt{car} task, visualized in Figure~\ref{fig:benchmarks}(l), we use the prompt \texttt{"Navigate to the red car."} In this case, there can be multiple valid reasoning traces. The Perception field could describe the scene as \texttt{"The red car is on the left of the gray car."} or equivalently as \texttt{"The gray car is on the right of the red car."} Both descriptions are correct.

However, hard-label imitation in standard SFT, as in Equation~\ref{eq:sft}, forces the model to match one particular description, even when several alternatives are valid. This may limit the model's ability to represent diverse but correct reasoning patterns. In contrast, our method learns from a distributional target, as in Equation~\ref{eq:fwdkl-grad-ours}, which can capture multiple valid perception modes and therefore provides more flexible supervision.
\section{Additional Training Details}\label{appendix: training_detail}

\subsection{Datasets} We use a mixture of data from the RECON~\cite{shah2021rapid} and FrodoBots-2K~\cite{frodobots2k} datasets for training and validation of \algname{}.
RECON is an outdoor visual navigation dataset collected with a Clearpath Jackal ground robot exploring an unstructured university-campus environment under diverse weather and lighting conditions. 
Each trajectory pairs a fisheye front-camera image stream with the robot's commanded actions, and covers scenes ranging from paved sidewalks and grassy fields to parking lots and pedestrian-shared spaces. 
FrodoBots-2K is a large-scale sidewalk navigation dataset collected with the Earth Rover Zero platform across more than 10 cities, providing thousands of hours of teleoperated driving in pedestrian-shared urban environments under varied lighting, weather, and crowd conditions. 
Due to the size of the dataset and infeasibility of full training in the project timeline, we choose to evaluate on a subset of the dataset consisting of 25\% of the scenes, randomly selected from the available online splits.
Compared to RECON's single-campus Jackal collection, FrodoBots-2K tests generalization on a substantially larger and more visually diverse corpus. The train/validation split is set to 90\%:10\%.

\subsection{Model and Training Setup}
We use Qwen2.5-VL-3B-Instruct~\cite{yang2024qwen2} as the backbone VLM for both the student and the self-teacher. The student is initialized from the released pretrained checkpoint, and the self-teacher is initialized with the same weights. During training, the self-teacher is updated as an exponential moving average of the student. We train \algname on a server with 8 A6000 Pro, 256G of RAM, AMD Threadripper 7980X CPU with 64 cores. The key hyperparameters used in our \algname are reported in Table~\ref{tab:hyperparameter}. 

\begin{table}[h]
\centering
\caption{\textbf{Training hyperparameters.} Main hyperparameters used for model training and inference in real-world experiments. Hyperparameters not listed here use the HuggingFace Transformers defaults.}
\label{tab:hyperparameter}
\footnotesize
\renewcommand{\arraystretch}{1.08}
\setlength{\tabcolsep}{4pt}

\resizebox{\textwidth}{!}{
\begin{tabular}{ll ll}
\toprule
\textbf{Hyperparameter} & \textbf{Value} &
\textbf{Hyperparameter} & \textbf{Value} \\
\midrule

Backbone VLM & Qwen2.5-VL-3B-Instruct &
Image resolution & dataset-native \\

Max output tokens & 512 &

Past waypoints & 8 \\

Future waypoints & 16 &Prediction horizon & 6.4 s \\
Waypoint frequency & 2.5 Hz & Train/validation split & 9:1 \\

Optimizer & adamw\_torch\_fused &
Learning rate & $1 \times 10^{-5}$ \\

Batch size per GPU & 1 &
Gradient accumulation & 2 \\

Training epochs & 2 &
LR schedule & cosine \\

Warmup steps & 5 &
Random seed & 42 \\

EMA rate $\alpha$ & 0.999 &
Teacher update frequency & every 10 steps \\

\bottomrule
\end{tabular}
}
\end{table}

\subsection{Full Prompt Templates for Training}
\label{appendix:prompts}

We show the complete prompt templates for the \texttt{FrodoBots-2K} dataset, which produces the paper's headline numbers.
All datasets follow the same structure; only environment-specific nouns (surface type, agent categories, action set) differ.
Goal-conditioned variants are marked \emph{(GC only)}.

\paragraph{Template variables.}
All query templates share the following format keys, populated per sample at training time.

\begin{center}
\footnotesize
\renewcommand{\arraystretch}{1.15}
\begin{tabular}{ll}
\toprule
\textbf{Variable} & \textbf{Description} \\
\midrule
\texttt{\{hist\_steps\}}  & Number of history waypoints shown (default: 8) \\
\texttt{\{hist\_hz\}}     & Effective history sample rate in Hz (default: 5\,Hz) \\
\texttt{\{hist\_text\}}   & Serialised list of past $[x, y]$ waypoints \\
\texttt{\{avg\_speed\}}   & Average speed over the history window (m/s) \\
\texttt{\{fut\_steps\}}   & Number of future waypoints to predict (default: 16) \\
\texttt{\{fut\_secs\}}    & Future prediction horizon in seconds (default: 6.4\,s) \\
\texttt{\{fut\_hz\}}      & Effective future sample rate in Hz (default: 2.5\,Hz) \\
\texttt{\{fut\_text\}}    & \emph{Teacher only:} ground-truth future trajectory JSON \\
\bottomrule
\end{tabular}
\end{center}

\paragraph{Self-Teacher and Student system prompt.}
Shared by all teacher template variants.

\begin{lstlisting}
You are an outdoor sidewalk-robot navigation planner.
You control an Earth Rover Zero navigating sidewalks and outdoor spaces.
Your sensor is a single forward-facing 1024x576 RGB camera mounted at
~0.45 m height.
Coordinates: x = forward, y = left (meters), origin at current position.
\end{lstlisting}

\paragraph{Teacher query template (4-section default).}
The assistant turn is pre-filled with \lstinline{Perception:}.

\begin{lstlisting}
Past {hist_steps} waypoints at {hist_hz} Hz
(last = [0.0, 0.0], avg speed {avg_speed} m/s):
{hist_text}

A human operator chose this path for the next {fut_secs} s
({fut_steps} points at {fut_hz} Hz).
Justify why this action is sensible using ONLY what is visible in the
underlying scene. Do not invent agents or obstacles that aren't actually
visible; many frames on this dataset are empty sidewalks, and "no agents"
is the correct answer whenever nothing is there.
Describe only the scene itself, not any markings or overlays on top of it.
A reader who sees only the raw image must be able to verify your reasoning.

Perception: 1-2 sentences with concrete visual detail about the sidewalk
and its surroundings -- path surface (concrete, brick, asphalt), edges
(curbs, grass, walls, parked cars), buildings or foliage, and how much of
the sidewalk is clear ahead.
Agents: list each agent actually visible in the image. If the scene is
empty, write exactly "none".
Prediction: If you wrote "none" above, write "none" here too. Otherwise
describe each listed agent's likely motion in 1-2 sentences.
Planning: Pick one action from [go straight, turn left, turn right,
slow down, stop, cross intersection, navigate around] and explain in 1-2
sentences why this is right, tying the choice to specific features you
just named in Perception (e.g. "the sidewalk runs straight past a row of
parked cars", "a driveway cuts across the path at about 6 m").

Trajectory: {"future_trajectory": {fut_text}}
\end{lstlisting}

\paragraph{Teacher Task directive \emph{(for Goal Conditioning)}.}
For goal conditioning, the following directive is inserted
immediately before \lstinline{Perception:} in the query above.
The assistant turn is pre-filled with \lstinline{Task:} instead of
\lstinline{Perception:}. The trainer extracts the emitted \lstinline{Task:}
sentence and injects it into the student assistant pre-fill; the student
never generates it.

\begin{lstlisting}
Task: one short sentence stating the high-level goal for this scene
(e.g., "navigate toward the building entrance", "navigate around the
tree", "navigate past the parked car"). Name a visible target object --
building, tree, gate, bench, parked car, sign, driveway, lamppost,
junction. Use positional descriptors ("on the left", "on the right")
when they help identify a specific object. Do not use ego-motion verbs
like "turn left" or "go straight" here -- those belong in Planning.
\end{lstlisting}


\textbf{Student query template.}

\begin{lstlisting}
Past {hist_steps} waypoints at {hist_hz} Hz
(last = [0.0, 0.0], avg speed {avg_speed} m/s):
{hist_text}

Predict the next {fut_steps} waypoints ({fut_secs} s at {fut_hz} Hz).

Answer in three short sentences, then output the trajectory JSON on a
single line.
Perception: sidewalk layout (1 short phrase) and agents from
[pedestrian, cyclist, vehicle, parked, obstacle, none].
Prediction: What will each agent do? Write "none" if no agents.
Planning: What action? Pick one from [go straight, turn left, turn right,
slow down, stop, cross intersection, navigate around] and give a short
reason.
Example: wide sidewalk, no agents; none; go straight, clear walkway.

Trajectory: {"future_trajectory": [[x1, y1], [x2, y2], ...]}
\end{lstlisting}

\paragraph{Student assistant pre-fill.}
The student's assistant turn is pre-filled as follows; loss is computed only
over the tokens that follow \lstinline{Perception:}.
The \lstinline{Task:} line \emph{(GC only)} is prompt context injected by the
trainer from the teacher's output --- the student never generates it.
 
\medskip
\noindent Default (non-GC):
\begin{lstlisting}
Perception:
\end{lstlisting}
 
\noindent Goal-conditioned \emph{(GC only)}, task sentence injected from teacher output:
\begin{lstlisting}
Task: <extracted task sentence>
Perception:
\end{lstlisting}

\textbf{KL Calculation.} Teacher\&student share the same reasoning format, except for a teacher-only inferred goal span, which is excluded when aligning response tokens. We compute Qwen's standard token-level KL divergence between teacher/student logits before token sampling as forward KL.

\subsection{Training code for reproducibility}
For reproducibility, we provide our training code in the Supplementary Material, including \texttt{data.py}, \texttt{prompt.py}, and \texttt{train.py}, together with a README file that describes how to run the training pipeline. Specifically, \texttt{data.py} constructs the image-and-text prompts and batches training samples, \texttt{prompt.py} defines the instruction templates for scene description and path justification, and \texttt{train.py} implements the two-pass self-distillation training loop.

\subsection{Inference Details}
We also summarize the goal-conditioned inference procedure in Algorithm~\ref{alg:asd-inference}.

\begin{algorithm}[h]
\caption{\algname Inference (Student Policy)}
\label{alg:asd-inference}
\KwIn{ New observation $o_0$, history $\tau_{-T:0}$, and language goal $g^\circledast$}
\KwOut{Future trajectory $\tau^\star_{0:H}$}
\While{the task is not terminated}{
    \nl $(\tilde{R}, \tilde{\tau}_{0:H}) \leftarrow \pi^\star_\theta(o_0, \tau_{-T:0}, g^\circledast)$, Eq. (7) \tcp*[r]{student only}
    Execute $\tilde{\tau}_{0:K},K\leq H$, using the robot controller\;
    Update $o_0$ and $\tau_{-T:0}$ from the latest robot state\;
}
\end{algorithm}


\begin{table}[h]
\centering
\caption{Prompt for language-conditioned navigation benchmarks.}
\renewcommand{\arraystretch}{1.15}
\addtolength{\tabcolsep}{-0.4em}
\begin{tabular}{cccl}
\toprule
 & \textbf{Task} & \textbf{Difficulty} & \textbf{Prompt} \\
\midrule

\multirow{3}{*}{Directions}
& Door        &  $\star$ & "Navigate straight to the door"\\
& left chair  &  $\star\star$ & "Navigate to the chair on the left"\\
& right chair &  $\star\star\star$ & "Navigate to the chair on the right"\\
\midrule

\multirow{4}{*}{Objects}
& Blue chair &  $\star$ & "Navigate to the blue chair"\\
& Trash Bin &  $\star$ & "Navigate to the trash bin"\\
& Chair  &  $\star\star$ & "Navigate to the chair"\\
& Monitor &  $\star\star\star$ & "Navigate to the monitor"\\
\midrule

\multirow{2}{*}{Multi-stage}
& Doorway &  $\star\star$ & "Navigate through the door"\\
& Hallway &  $\star\star\star$ & "Navigate through the door, turn right, then follow the hallway"\\
\midrule

\multirow{3}{*}{Outdoor}
& Sidewalk  & $\star$ & "Navigate along the sidewalk"\\
& Statue & $\star\star$ & "Navigate to the statue"\\
& Tree  & $\star\star$ & "Navigate to the tree"\\
& Car & $\star\star\star$ & "Navigate to the red/gray car"\\
\bottomrule
\end{tabular}
\label{tab:prompt}
\end{table}

\section{Language-conditioned Navigation Tasks}\label{appendix:real_world_task}
\begin{figure}[t]
    \centering
    \includegraphics[width=1\linewidth]{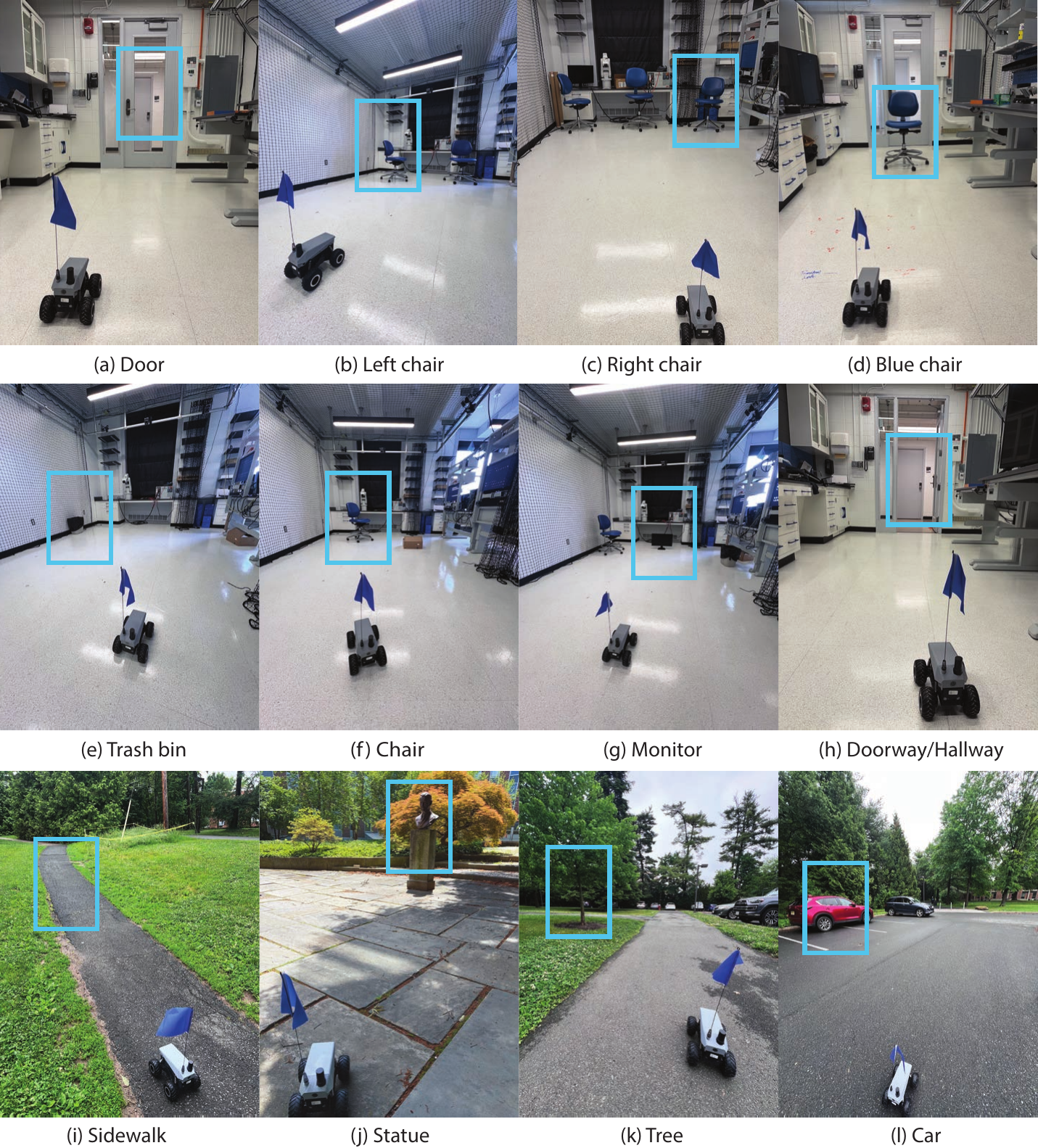}
    \caption{\textbf{Summary of Benchmark Tasks.} Panels (a)--(h) show the indoor tasks reported in Table~1, covering directional, object-centric, and multi-stage navigation scenarios. Panels (i)--(l) show the outdoor tasks reported in Table~2. The target object or goal region for each task is highlighted with a cyan bounding box.}
    \label{fig:benchmarks}
\end{figure}

 Here we provide details of the tasks reported in Section 4. All indoor tasks are evaluated in a laboratory environment containing typical office items such as chairs, trash bins, boxes, and monitors. All outdoor tasks are evaluated in a campus environment. Figure~\ref{fig:benchmarks} summarizes all the testing cases, and Table~\ref{tab:prompt} reports the prompt used in each benchmark, which is fixed across \algname, OmniVLA and NaVILA.

\subsection{Indoor Task Setting}\label{appendix:indoor_task}
\subsubsection{Directions} We evaluate the model’s ability to follow directional instructions when selecting among different target objects. The tasks are ordered from easiest to hardest: \texttt{door}, \texttt{left chair}, and \texttt{right chair}. In the \texttt{door} task (a), the robot is placed in front of a door with no obstacles between them. The user prompt for this task is "\texttt{Navigate straight to the door}", which tests \algname{}'s capability to go straight. In the \texttt{left chair} task (b), two identical chairs are placed in front of the robot, side by side, and the robot must navigate to the left chair, by following the user prompt: "\texttt{Navigate to the chair on the left.}" In the \texttt{right chair} task (c), three identical chairs are placed side by side, and the robot must navigate to the right chair, as following the task prompt: "\texttt{Navigate to the chair on the right.}" The \texttt{door}, \texttt{left chair}, and \texttt{right chair} tasks are visualized in Figure~\ref{fig:benchmarks}(a)--(c), respectively. This set of experiments test the directional steering capability, of which the task prompts can be found in Table~\ref{tab:prompt}.

\subsubsection{Objects} We evaluate the model's ability to distinguish different objects. The tasks are ordered from easiest to hardest: \texttt{blue chair}, \texttt{trash bin}, \texttt{chair} and \texttt{monitor}. In the \texttt{blue chair} task, a blue chair is placed within the robot's field of view, and the robot is prompted to \texttt{"Navigate to the blue chair"}. In the \texttt{trash bin} task, a trash bin is placed in a similar setting, and the robot is prompted to \texttt{"Navigate to the trash bin"}. In the \texttt{chair} task, a trash bin and a chair are placed in front of the robot, and the robot is prompted to \texttt{"Navigate to the chair"}. In the \texttt{monitor} task, a trash bin, a chair, and a monitor are placed in the scene, and the robot is prompted to \texttt{"Navigate to the monitor"}. These four tasks are visualized in Figure~\ref{fig:benchmarks}(d)--(g), and the corresponding prompts are summarized in Table~\ref{tab:prompt}.

\subsubsection{Multi-stage} We evaluate the model’s ability to complete multi-stage tasks that require executing multiple navigation behaviors in sequence. The tasks are \texttt{doorway} and \texttt{hallway}. In the \texttt{doorway} task, the robot is placed in front of a door and is prompted to \texttt{"Navigate through the door"}. In the \texttt{hallway} task, the robot starts from the same physical setup, but must pass through the door, turn right, and then move along the hallway. The corresponding prompt is \texttt{"Navigate through the door, turn right, then follow the hallway"}. Both tasks are visualized in Figure~\ref{fig:benchmarks}(h). They share the same physical setup but require different navigation behaviors. The prompts are summarized in Table~\ref{tab:prompt}.

\subsection{Outdoor Tasks Setting}\label{appendix:outdoor_task}
We evaluate the model's ability to navigate to different target objects in outdoor environments, including four tasks: \texttt{sidewalk}, \texttt{statue}, \texttt{tree}, and \texttt{car}. In the \texttt{sidewalk} task (i), the robot navigates along a sidewalk running through the middle of a grassy area, following the prompt \texttt{"Navigate along the sidewalk"}. In the \texttt{statue} task (j), the robot navigates to a bronze statue, following the prompt \texttt{"Navigate to the statue"}. In the \texttt{tree} task (k), the robot is placed on a sidewalk and needs to navigate to a tree by crossing a grassy area, following the prompt \texttt{"Navigate to the tree"}. In the \texttt{car} task (l), the robot starts in a garage and navigates to a car of a specific color, following the prompt \texttt{"Navigate to the red/grey car"}. These tasks are visualized in Figure~\ref{fig:benchmarks}(i)--(l), and the corresponding prompts are summarized in Table~\ref{tab:prompt}.

\subsection{Evaluation Fairness, Execution and Stopping Criteria}
\textbf{Evaluation Fairness and Role of Reasoning.} Across 10 trials per task, evaluators did \emph{not} provide corrective interventions during execution (e.g. repositioning or issuing recovery commands); so, mean steps to corrective intervention is not applicable. 
During planning, we evaluate the safety and reasonability of the predicted trajectory, \& give all methods three times to replan per episode (same protocol for all models). Episode fails if the robot collided with an obstacle or if the goal is no longer in its FOV.

\textbf{Execution Behavior and Stopping.} Similar to OmniVLA in evaluation/stopping criteria, task completion is judged by method-blinded human evaluator, who stops the episode once the instruction is satisfied. \algname{} has no learned stopping criterion. Pauses in multi-stage demonstrations are not emergent; they reflect observation and inference latency as \algname replans after executing 8 of 16 predicted actions. OmniVLA predicts only 8 actions, all of which are executed for fairness.

\subsection{Real World Experiments Baselines Implementation}

\subsubsection{OmniVLA} 
We evaluate OmniVLA~\citep{hirose2025omnivla}, a vision-based navigation policy that takes a language prompt as the goal input. We use the pretrained \textit{omnivla-original} checkpoint. During the inference, OmniVLA is queried with the current front-view image and the goal text, and outputs a sequence of 8 future actions.
Before execution, the predicted action sequence is checked by human intuitive judgment to determine whether it is safe and reasonable. If the plan is considered unreasonable, OmniVLA is required to replan. We allow at most 3 such heuristic interventions per run; once this limit is exceeded, the run is marked as a failure. For trajectory continuity and smoothness, all eight predicted actions are executed before the next front-view image is acquired. Each run is terminated when the human evaluator judges the task to be successful, or when a collision occurs.

\subsubsection{NaVILA}
We also evaluate NaVILA~\cite{cheng2024navila}, a two-level navigation framework that combines a vision-language-action model with low-level locomotion skills. We use the pretrained NaVILA's VLA checkpoint to generate high-level language commands and execute them using the built-in controller of the FrodoBot. At each planning step, the model predicts a future action command, such as ``forward 75 cm'' or ``turn right 45$^\circ$''. The robot then executes these commands before receiving a new plan.
For the multi-stage tasks, \texttt{Doorway} and \texttt{Hallway}, we use the 64-frame checkpoint, which conditions on the previous 64 front-camera frames. For the remaining tasks, we use the 8-frame checkpoint. As in the OmniVLA evaluation, we allow at most three heuristic interventions per run; once this limit is exceeded, the run is marked as a failure. Each run terminates when the human evaluator judges the task to be successful, or when a collision occurs.

\subsubsection{Vanilla Qwen2.5-VL} We perform inference by replacing the \algname checkpoint with the vanilla Qwen2.5-VL-3B-Instruct model, while keeping all other inference settings unchanged.

\subsubsection{Claude Opus 4.7} Claude Opus 4.7 is run in auto mode with low reasoning effort. At each inference step, Claude receives the current front-view image, reasons about the navigation goal, and generates executable motion commands, including the duration for turning and moving forward. The robot then executes the generated command before the next inference step. We choose low reasoning effort because Claude’s reasoning latency is very high, and using low effort reduces the inference time. 
We use two helper functions, \texttt{save_fig.py} and \texttt{send_command.py}, to interact with the robot. The function \texttt{save_fig.py} saves the current front-view image from the robot to disk, while \texttt{send_command.py} sends the generated control command to the robot. We then prompt Claude to navigate the robot using the following instruction:
\begin{lstlisting}[escapechar=!]
Task: {task-specific prompt listed in Table 6}.
Please read the current front-view image saved by save_fig.py. Based on this image and the task goal, generate a control command and send it to the robot using send_command.py. After the command is executed, call save_fig.py again to obtain the updated front-view image. Repeat this observe-plan-act loop until the task is completed. Stop the process when you judge that the robot has reached the target.
\end{lstlisting}

\section{Additional Real World Qualitative and Open-Loop Quantitative Results}

\subsection{Open-Loop Evaluation}
We conduct the open-loop experiments using the non-goal-conditioned \algname model, where no language goal is provided to the student. The purpose of this comparison is to evaluate \algname{} in the open-loop trajectory prediction setting and is the best among competitive supervised finetuning baselines, including Gaussian Policy, Gaussian Mixture model, Diffusion Policy, Qwen2.5-VL + SFT, Alpamayo-1+SFT and OpenVLA+SFT.


\subsection{Open-Loop Experiment Setup}

In addition to the real-world language-conditioned navigation tasks reported in Section 4, we conduct an open-loop evaluation on the validation splits of RECON and FrodoBots-2K. Following the inference formulation in Equation (7), the model receives the current front-view image and the past ego-frame trajectory as input, and predicts a future trajectory without closed-loop execution. We train our \algname with 8 past waypoints as history and predict of future waypoints, corresponding to a 6.4-second horizon at 2.5 Hz. The predicted trajectory is compared directly with the ground-truth future trajectory. All baselines are trained and evaluated on the same train/validation split.


\subsection{Open-Loop Baselines}
We compare against two groups of baselines. The first group consists of non-VLA policies trained directly on the same data, including Gaussian Policy, Gaussian Mixture Model~\citep{reynolds2009gaussian}, and Diffusion Policy~\citep{chi2025diffusion}. The second group consists of VLA backbones fine-tuned with standard SFT, including Qwen2.5-VL~\cite{yang2024qwen2}, Alpamayo~\citep{wang2025Alpamayo}, and OpenVLA~\citep{kim2024openvla}. The training details of each baseline are:

\paragraph{Gaussian Policy.}
We adopted the robomimic's built-in \textbf{BC\_Gaussian} algorithm~\citep{robomimic2021}. Robomimic is a framework for robot learning from demonstration, which offers a broad set of offline learning algorithms to learn from datasets. The current front image is encoded by a ResNet18 with spatial softmax pooling, then linearly projected into a flat feature vector. The 8 previous ego-frame waypoints are flattened into a 16 dimensional vector. We concatenate the visual vector and position vector into a fusion vector, then pass it through a shared two-layer MLP, after which two parallel MLPs predict the mean and standard deviation of a single 32-dimensional diagonal Gaussian that jointly parameterized the 16 future waypoints. The training objective is the negative log-likelihood of the ground-truth action under this distribution. All the training hyperparameters are kept at robomimic defaults. At evaluation we deterministically output the predicted Gaussian mean.

\paragraph{Gaussian Mixture Model.}
We adopted robomimic's built-in \textbf{BC\_GMM} algorithm. The observation encoder, history fusion, and shared MLP trunk are identical to the Gaussian policy; the only architectural difference is the policy head, which parametrizes a 5-mode mixture of 32-dimensional diagonal Gaussians instead of a single one. Training uses the negative log-likelihood of the ground-truth action under the mixture density. At evaluation we draw one sample from the predicted mixture.

\paragraph{Diffusion Policy.}
We implement the Diffusion Policy baseline on top of the official codebase~\citep{chi2025diffusion}. Diffusion Policy was originally proposed for robot manipulation, where a DDPM~\citep{ho2020denoising} iteratively denoises Gaussian noise into an end-effector trajectory conditioned on visual observations. The observation encoder and history fusion are identical to the BC\_Gaussian and BC\_GMM baselines (ResNet18 + spatial-softmax pooling, concatenated with the 16-d history vector). As for the policy, we use a conditional U-Net that convolves along the temporal dimension to predict the noise given the fused observation. The network is trained to predict the added noise. At evaluation we sample one noise and denoise it to get the trajectory.

\paragraph{Qwen+SFT.} This is implemented as \algname{}'s student policy training.

\paragraph{Alpamayo-1+SFT.}

We evaluate Alpamayo-R1~\citep{wang2025Alpamayo}, a vision-language trajectory model that pairs a frozen Qwen3-VL-8B visual encoder with a small action-expert denoiser that generates the future trajectory via diffusion-style sampling. We fine-tune the expert denoiser (Stage 2 only, VLM frozen) on RECON and FrodoBots-2K for 5 epochs at batch size 8 and learning rate 1e-4. At inference, the model draws a single trajectory sample per query (num-traj=1) for fair single-prediction comparison against the other baselines.

\paragraph{OpenVLA+SFT.}

We evaluate OpenVLA~\citep{kim2024openvla}, a 7B-parameter Prismatic vision-language-action model that emits a single discretized \texttt{(v\_lin, v\_ang)} action token pair per image, which we then integrate through a unicycle model to obtain the predicted trajectory. We add a LoRA adapter of rank 32 and fine-tune on RECON and FrodoBots-2K for 5 epochs at batch size 32 and learning rate 5e-4, deriving action targets from each sample's first ground-truth future step.


\subsection{Open-Loop Quantitative Evaluation.}
Table~\ref{tab:open_loop_results} summarizes 2-second open-loop trajectory prediction results on RECON and FrodoBots-2K,  where we report the Absolute Trajectory Error (ATE), Average Displacement Error (ADE), Mean Absolute Error (MAE) and Mean Squared Error (MSE). \algname achieves the lowest error on 7 out of 8 comparisons. \algname has a significant improvement over its direct baseline Qwen+SFT. Since both models share the same Qwen2.5-VL backbone, training data, and optimization, the gap is attributable to the reasoning-token supervision rather than to capacity or training recipe. \algname also uniformly surpasses non-VLA behavior cloning methods, indicating that the gain is not merely closing a VLA-to-specialist gap. The remaining VLA baselines, Alpamayo-1 and OpenVLA, fall well behind under SFT alone, suggesting that the choice of pretrained backbone is itself a strong factor and that navigation behavior does not transfer automatically from any driving VLA. Based on these results, we conclude that the feedback-conditioned reasoning distillation improves short-horizon trajectory prediction over standard SFT, and the gain persists at a larger and more diverse data scale.

\begin{table*}[h]
\centering
\caption{
Open-loop evaluation on RECON and FrodoBots-2K. \algname achieves the best performance across both datasets under the open-loop evaluation protocol. Lower is better for all metrics.
}
\label{tab:open_loop_results}
\resizebox{\textwidth}{!}{
\begin{tabular}{lcccccccc}
\toprule
\multirow{2}{*}{Method}
& \multicolumn{4}{c}{RECON} 
& \multicolumn{4}{c}{FrodoBots-2K} \\
\cmidrule(lr){2-5} \cmidrule(lr){6-9}
& ADE $\downarrow$ & ATE $\downarrow$ & MAE $\downarrow$ & MSE $\downarrow$
& ADE $\downarrow$ & ATE $\downarrow$ & MAE $\downarrow$ & MSE $\downarrow$ \\
\midrule
Gaussian Policy 
& 0.226 & 0.321 & 0.144 & 0.059
& 0.034 & 0.046 & 0.019 & \textbf{0.002} \\

Gaussian Mixture Model  
& 0.260 & 0.364 & 0.166 & 0.084
& 0.036 & 0.050 & 0.020 & 0.002 \\

Diffusion Policy~\cite{chi2025diffusion}
& 0.180 & 0.238 & 0.108 & 0.045
& 0.034 & 0.045 & 0.020 & 0.003 \\

Qwen2.5-VL+SFT~\cite{yang2024qwen2}
& 0.192 & 0.247 & 0.115 & 0.042
& 0.064 & 0.061 & 0.027 & 0.007 \\

Alpamayo-1+SFT~\cite{wang2025Alpamayo}
& 0.303 & 0.362 & 0.188 & 0.091
& 0.093 & 0.106 & 0.055 & 0.041 \\

OpenVLA+SFT~\cite{kim2024openvla}
& 0.341 & 0.417 & 0.208 & 0.114
& 0.246 & 0.269 & 0.138 & 0.072 \\
 
\midrule
\textbf{Qwen2.5-VL-\algname} & \textbf{0.174} & \textbf{0.227} & \textbf{0.104} & \textbf{0.037}
& \textbf{0.031} & \textbf{0.042} & \textbf{0.017} & 0.004 \\
\bottomrule
\end{tabular}
}
\end{table*}


\subsection{Real World Experiments Qualitative Analysis}
In this section, we analyze the navigation performance beyond the success rate reported in our Table 1 and Table 2 of the main text. We visualize representative examples from both indoor and outdoor settings and compare them with OmniVLA~\cite{hirose2025omnivla} and NaVILA~\cite{cheng2024navila}. We observe consistent performance improvements over OmniVLA and NaVILA across several evaluation settings.

\begin{figure}[h]
    \centering
    \includegraphics[width=\linewidth]{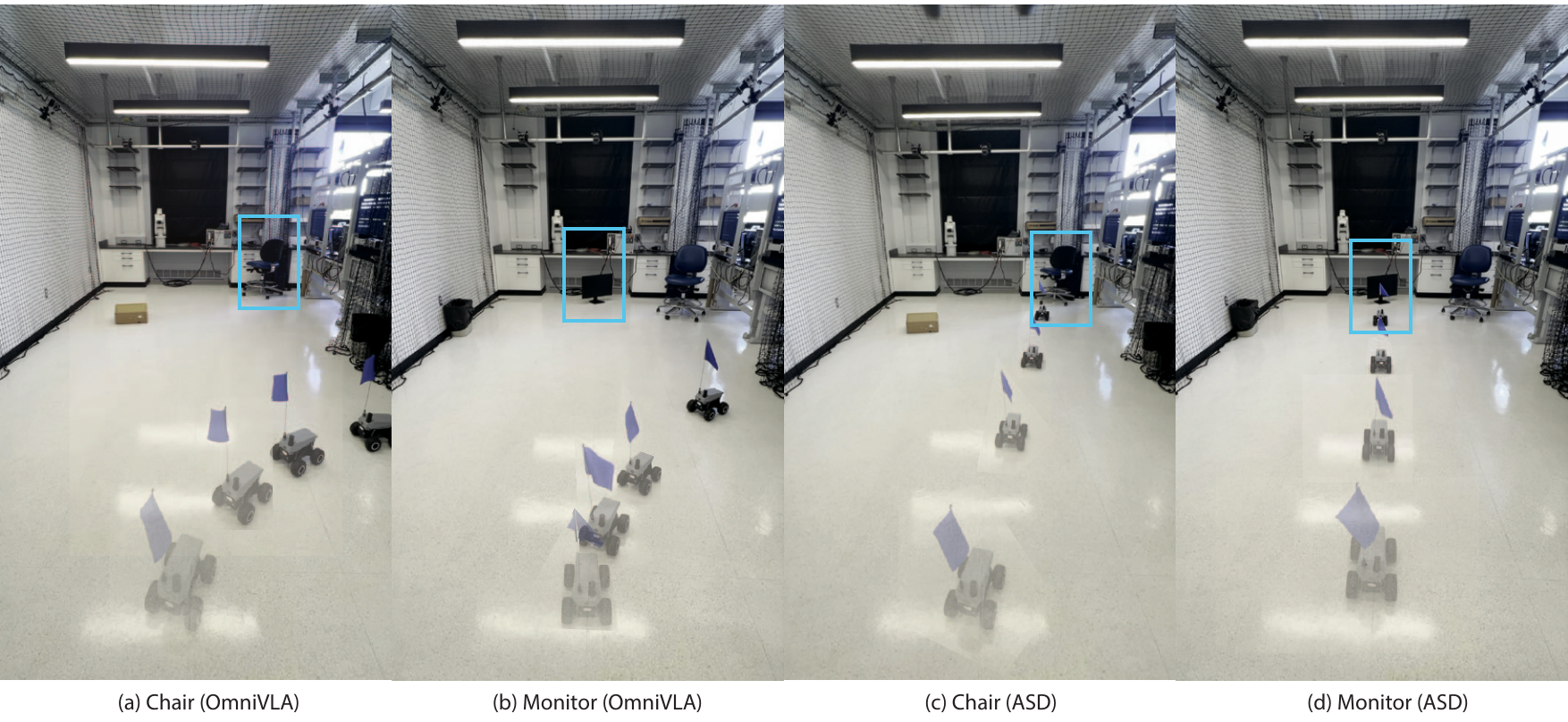}
    \caption{\textbf{Evaluation on  Multi-object Tasks.} We visualize the \texttt{Chair} and \texttt{Monitor} tasks for OmniVLA (left) and \algname (right). \textbf{(a)} OmniVLA navigates the robot toward a distractor object, the net, instead of the target chair. \textbf{(b)} OmniVLA again navigates toward the net instead of the target monitor. \textbf{(c)} \algname correctly directs the robot to the chair. \textbf{(d)} \algname correctly directs the robot to the monitor.}
    \label{fig:object}
\end{figure}

\paragraph{Improved performance in multi-object tasks}
We observe improved performance of \algname in multi-object tasks, such as \texttt{Chair} and \texttt{Monitor} in the indoor setting. In these tasks, OmniVLA achieves success rates of $40\%$ and $20\%$, respectively, while our \algname reaches $90\%$ and $70\%$. This suggests that our method is better able to distinguish the target object from multiple candidate objects in the scene. Figure~\ref{fig:object} visualizes representative examples comparing OmniVLA and \algname. OmniVLA navigates the robot toward a distractor object (i.e., net) instead of the target (Figure~\ref{fig:object}a and Figure~\ref{fig:object}b), while \algname correctly directs the robot to the specified target (Figure~\ref{fig:object}c and Figure~\ref{fig:object}d).

\begin{figure}[b]
    \centering
    \includegraphics[width=1\linewidth]{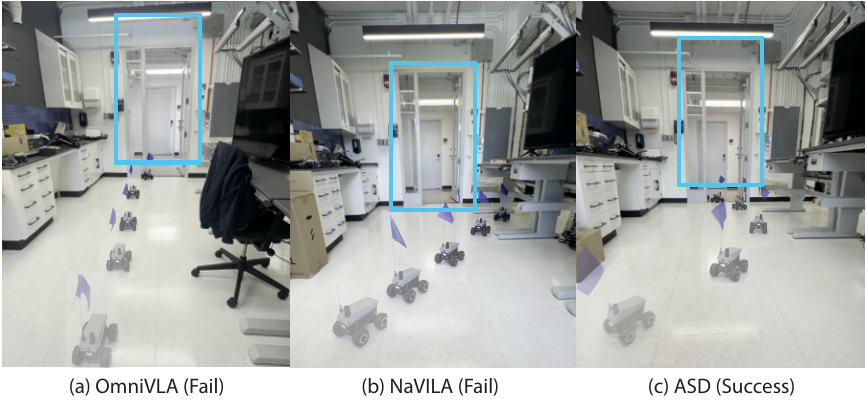}
    \caption{\textbf{Evaluation on \texttt{Doorway} tasks.} We visualize \algname's performance on the \texttt{Doorway} task compared with OmniVLA and NaVILA. \textbf{(a)} OmniVLA drives the robot straight forward and collides with the door frame on the left corner. \textbf{(b)} NaVILA also drives the robot straight forward and collides with the table on the right. \textbf{(c)} \algname produces a left-turning trajectory that allows the robot to pass through the door successfully.}
    \label{fig:stage}
\end{figure}

\begin{figure}
    \centering
    \includegraphics[width=1\linewidth]{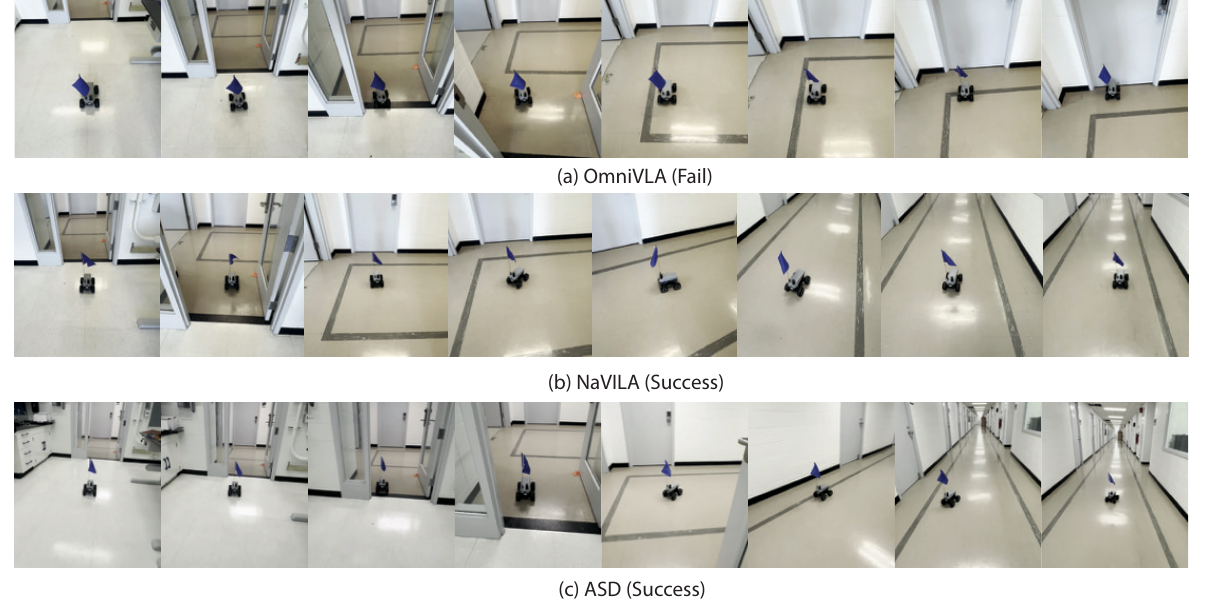}
    \caption{\textbf{Evaluation on \texttt{Hallway} tasks.} We visualize \algname's performance on the \texttt{Hallway} task compared with OmniVLA and NaVILA. \textbf{(a)} OmniVLA navigates the robot through the door but fails to complete the subsequent right turn and hallway-following behavior. \textbf{(b)} NaVILA successfully completes the full multi-stage instruction. \textbf{(c)} \algname also successfully completes the full multi-stage instruction.}
    \label{fig:stage2}
\end{figure}

\paragraph{Generalization to multi-stage tasks}
We observe that \algname generalizes well to multi-stage tasks, such as \texttt{Doorway} and \texttt{Hallway}. None of the baselines succeeds reliably on both tasks: OmniVLA achieves a $70\%$ success rate on the two-stage \texttt{Doorway} task but fails on the four-stage \texttt{Hallway} task ($0\%$), while NaVILA achieves a $50\%$ success rate on \texttt{Hallway} but performs poorly on \texttt{Doorway} ($30\%$). In contrast, our \algname achieves higher success rates on both tasks, reaching $80\%$ and $60\%$, respectively. Figure~\ref{fig:stage} visualizes representative comparisons on these tasks. In the \texttt{Doorway} task, NaVILA drives the robot straight forward and collides with the table (Figure~\ref{fig:stage}b), while \algname produces a left-turning trajectory that allows the robot to pass through the door (Figure~\ref{fig:stage}c). In the \texttt{Hallway} task, OmniVLA only navigates the robot through the door and fails to complete the subsequent right turn and hallway-following behavior (Figure~\ref{fig:stage2}a). In contrast, \algname successfully completes the full multi-stage instruction (Figure~\ref{fig:stage2}c).

\begin{figure}
    \centering
    \includegraphics[width=1\linewidth]{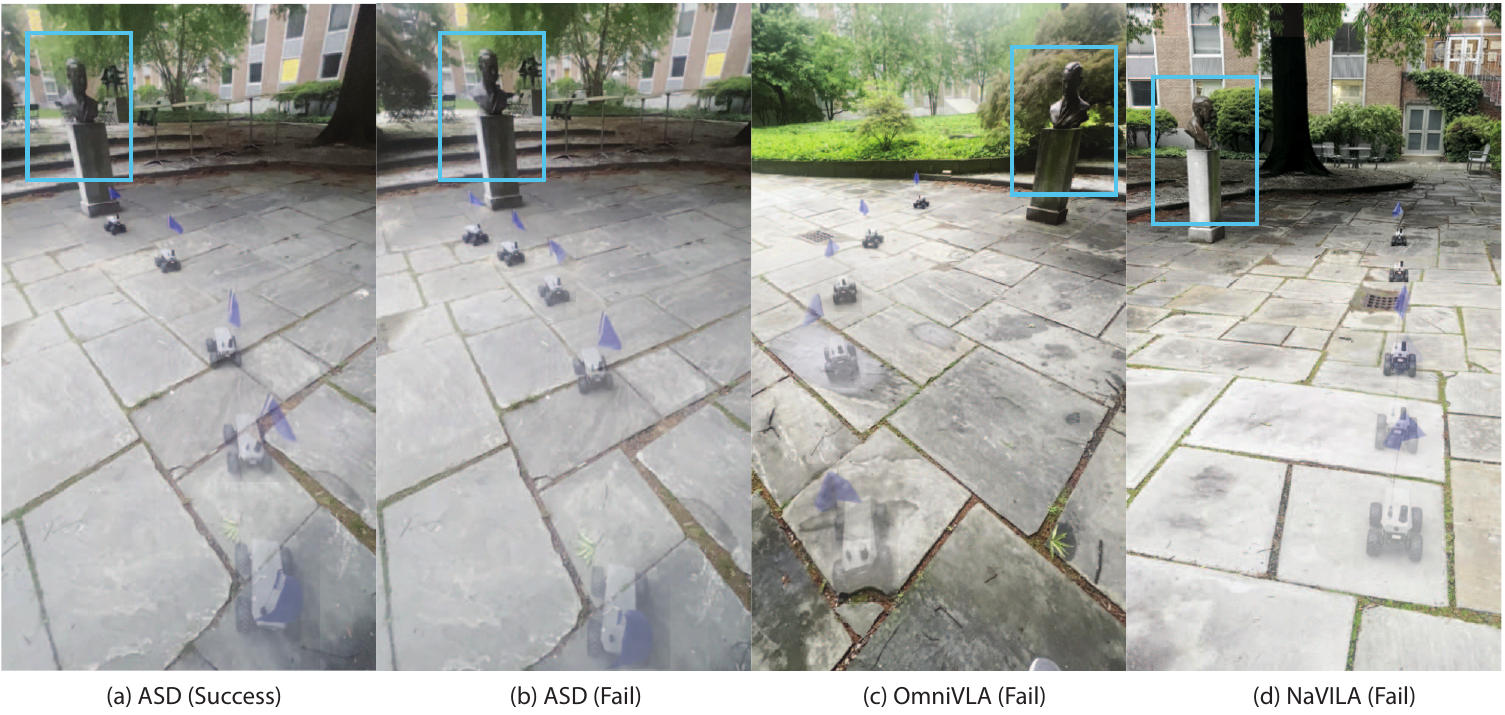}
    \caption{\textbf{Generalization to Bronze Statue. }We evaluate \algname, OmniVLA, and NaVILA on the \texttt{Statue} task and visualize representative examples. \textbf{(a)} \algname successfully reaches the target statue. \textbf{(b)} \algname produces a correct plan but fails to reach the target due to execution error on uneven terrain. \textbf{(c)} OmniVLA fails to generate a suitable trajectory toward the statue. \textbf{(d)} NaVILA moves mostly straight and fails to reach the target, due to a mismatch between the prompt style and the model’s expected instruction format.}
    \label{fig:statue}
\end{figure}

\paragraph{Generalization to out-of-distribution objects.} We observe stronger performance on the \texttt{Statue} task, where the bronze statue can be regarded as an out-of-distribution object with respect to our training datasets. Despite this OOD challenge, our \algname achieves an $80\%$ success rate, outperforming OmniVLA ($60\%$) and NaVILA ($10\%$). Figure~\ref{fig:statue} visualizes representative comparisons among \algname, OmniVLA, and NaVILA on the \texttt{Statue} task. OmniVLA fails to generate a suitable trajectory toward the statue (Figure~\ref{fig:statue}c), and NaVILA moves mostly straight and fails to reach the target (Figure~\ref{fig:statue}d). The main failure mode of \algname, illustrated Figure~\ref{fig:statue}b, is related to execution: \algname predicts and executes the next 8 steps at once, so when the terrain is not sufficiently flat, the robot may deviate from the planned trajectory during execution.

\paragraph{Model Collapse and Self-Teacher Reasoning Drift.}\textit{(1) Self-teacher stays close to the non-fine-tuned Qwen checkpoint.}  With $\alpha=0.999$ and one EMA update every 10 training steps, approximately $93\%$ of the teacher's EMA weight remains assigned to the pretrained checkpoint, and $7\%$ stems from subsequent student updates. 
\textit{(2)  Even the most-changed tokens remain stable.} In a more aggressive stress test ($\alpha=0.99$, updated every step), $\mathrm{KL}(\mathrm{Qwen}\|\text{our ckpt})$ increases smoothly from $0.34$ to $0.68$ nats/token across 500 scenes, without abrupt changes. The most sensitive tail also changes only modestly: p95 surprisal increases from $4.8$ to $6.7$, while the maximum just increases from $16.8$ to $18.0$ nats, indicating proper amount task adaption rather than collapse. 
\textit{(3) Reasoning remains neither repetitive nor random.} Collapse may move reasoning-token entropy toward two extremes: zero for repetitive sentences, or the uniform limit ($\approx12.4$ nats) for random texts.
Our entropy shifts mildly ($1.36\to1.67$ nats); it neither suffers from the repetitive reasoning nor random text.

\section{Video Narratives}\label{appendix: video_text}
We provide a comprehensive video on project website that describes the core motivation and methodology behind \algname{}. This video features physical real-world demonstrations along with key qualitative comparisons highlighting tasks where our method significantly outperforms the baselines. In the side-by-side video comparisons, a green border denotes a successful trial, while a red border indicates a navigation failure. Notably, we also detail the inference time efficiency of each approach: while \algname{} runs entirely at normal speed without requiring any acceleration, the leading baselines—including OmniVLA and NaVILA—require substantial video speed-ups (e.g., \(4\times\), \(6\times\), or \(8\times\)) to match our real-time execution.

\subsection{Multi-Stage Task}
We first show multi-stage demos: \texttt{Navigate through the door, turn right and then follow the hallway (Hallway)}; \texttt{Navigate right to the sidewalk and navigate along the sidewalk.} Those demos are impressive as \algname{} can generalize decently well to multi-stage task prompts with the training only on the single-stage tasks.

\algname{} demonstrates strong generalization, achieving a \(60\%\) success rate on multi-stage Hallway navigation tasks despite training only on single-stage tasks, outperforming the \(50\%\) rate of the NaVILA baseline. While NaVILA is adept at multi-stage goals, it shows limited task generalization, with success rates of only \(27\%\) indoors and \(33\%\) outdoors on single-task prompts, as detailed in Tables 1 and 2.

We further evaluate \algname{} on a more complex, outdoor multi-stage task to demonstrate its advanced contextual reasoning capabilities. In this scenario, the agent is presented with a choice between two distinct sidewalks. The multi-stage prompt instructs the agent to: \(\texttt{"Navigate\ right\ to\ the\ sidewalk\ and\ navigate\ along\ the\ sidewalk."}\) To succeed, \algname{} cannot rely on simple forward movement; it must first spatially reason about its environment to differentiate between the sidewalk directly ahead and the one situated to its right. It then must successfully execute the sequence of navigating to and following the correct path. This demo highlights the model's ability to handle fine-grained spatial ambiguity in unconstrained outdoor environments.

\subsection{Extreme OOD Task}
We further demonstrate \algname{}'s generalization capability by evaluating it on a single-task goal prompt featuring an extreme out-of-distribution (OOD) object. Specifically, the task requires the agent to \(\texttt{"Navigate\ to\ the\ statue,"}\) a target object entirely absent from the training dataset. \algname{} achieves an impressive \(80\%\) success rate on this task. To ensure the robustness of this evaluation, we vary the initial starting positions, forcing the agent to execute specific left or right steering behavior to reach the target. This variation provides strong evidence that the agent's success is not a byproduct of a blind directional bias, but rather stems from its deep semantic understanding and precise goal-directed steerability. This task is from main Table 2 called \texttt{statue}, where OmniVLA achieves $60\%$ and NaVILA achieves $10\%$ success rate.

\subsection{Indoor Tasks}
We present video demonstrations comparing \algname{} against leading baselines on medium- and hard-difficulty tasks, highlighting instances of successful navigation by our method alongside baseline failures. 

In the hard mode's directional capability task, \(\texttt{"Navigate\ to\ the\ right\ chair,"}\) the agent must select the correct target from among three identical chairs placed in front of it. On this task, \algname{} achieves an \(80\%\) success rate, significantly outperforming OmniVLA at \(20\%\) and NaVILA at \(10\%\). The video qualitative results show that the strongest baseline, OmniVLA, consistently selects the incorrect direction—steering toward the left chair instead of navigating to the rightmost option. This failure mode shows our method's better directional semantic understanding during goal-directed navigation.

In the medium-difficulty object selection task, the agent is presented with two objects in front of it: a box and a chair. Given the target prompt, \(\texttt{"Navigate\ to\ the\ chair,"}\) \algname{} achieves a \(90\%\) success rate, substantially outperforming the leading baseline, OmniVLA, which achieves only \(40\%\). The video qualitative results show that OmniVLA consistently deviates to the black net on the right. In contrast, \algname{} demonstrates an efficient, direct, and clean trajectory toward the target object.

In the hard-difficulty object selection task, the agent is presented with three distinct objects: a trash can, a monitor, and a chair. When evaluated on the target prompt, \(\texttt{"Navigate\ to\ the\ monitor,"}\) \algname{} achieves a \(70\%\) success rate, significantly outperforming OmniVLA's \(20\%\) success rate. Consistent with the trends observed in the medium-difficulty task, the qualitative video results show that OmniVLA often deviates toward the black net on the right. In contrast, \algname{} exhibits a much more direct trajectory toward the correct target.

In the long-horizon single-stage evaluation, the prompt \(\texttt{"Navigate\ through\ the\ door"}\) explicitly tests the model's long-horizon planning and execution capabilities. The qualitative video results demonstrate that \algname{} outperforms the leading baseline, OmniVLA, by a margin of \(10\%\). This performance gap stems from a distinct failure mode in the baseline: OmniVLA crashes into the door frame, whereas \algname{} successfully navigates through the narrow opening without any collisions.

In the multi-stage task, \texttt{"Navigate\ through\ the\ door,\ turn\ right,\ then\ follow\ the\ hallway,"} the leading baseline, NaVILA, performs comparably to our method, with \algname{} outperforming it by a margin of \(10\%\). Notably, while the video demonstrations show that both models achieve successful navigation, \algname{} generates a significantly smoother trajectory during turns. Furthermore, our model achieves this performance running at normal speed, whereas the NaVILA baseline requires a \(4.7\times\) speed-up to execute the task efficiently.

\subsection{Outdoor Tasks}

We present video comparisons of \algname{} against leading baselines on the \texttt{Statue}, \texttt{Tree}, and \texttt{Car} tasks. These video qualitative results contrast the typical failure modes of the baseline models with the successful execution of our method. 

For instance, in the \(\texttt{"Navigate\ to\ the\ Statue"}\) task, OmniVLA correctly identifies the target direction but fails to navigate sufficiently close to the statue itself. This precision error highlights the origin of our model's \(20\%\) success rate advantage over the baseline.

In both the \(\texttt{Tree}\) and \(\texttt{Car}\) tasks, both \algname{} and the leading baseline, OmniVLA, achieve a perfect success rate of \(100\%\). However, OmniVLA suffers from extremely slow inference times compared to \algname{}. Specifically, OmniVLA requires a \(8.6\times\) and \(6.8\times\) speed-up on the \(\texttt{Tree}\) and \(\texttt{Car}\) tasks, respectively, to match the execution efficiency of our method.
\end{document}